\documentclass{article}

 \usepackage[preprint]{neurips_2026}

\usepackage[utf8]{inputenc} 
\usepackage[T1]{fontenc}    
\usepackage{hyperref}       
\usepackage{url}            
\usepackage{booktabs}       
\usepackage{amsfonts}       
\usepackage{nicefrac}       
\usepackage{microtype}      
\usepackage{xcolor}         

\usepackage[dvipsnames]{xcolor}
\usepackage[table]{xcolor}

\usepackage{enumitem}
\usepackage{amsmath}
\usepackage{amsthm}
\usepackage{subcaption}
\usepackage{cleveref}
\usepackage{graphicx}
\usepackage{newfloat}
\usepackage{colortbl}
\usepackage{multirow, multicol}
\usepackage{pifont}

\usepackage{arydshln}
\usepackage{algorithm}
\usepackage{algpseudocode}
\usepackage[normalem]{ulem}

\title{
\textcolor{gray}{
    \sout{Elastic-DETR: Making Image Resolution Learnable} \\ \sout{with Content-Specific Network Prediction} \\
}
Visual Accommodation: Rethinking Image Scale \\ as a Learnable Variable for Object Detection
}

\definecolor{UltraLightGray}{gray}{0.975}
\definecolor{VeryLightGray}{gray}{0.925}
\definecolor{LightGray}{gray}{0.875}
\definecolor{DarkGray}{gray}{0.825}

\author{%
  Daeun Seo$^1$\quad
  Heoseok Yang$^2$\quad
  Sihyeong Park$^3$\quad
  Hyungshin Kim$^1$ \\
  $^1$Chungnam National University\quad
  $^2$Santa Clara University\quad
  $^3$Korea Electronics Technology Institute \\
  \footnotesize\texttt{daeun@g.cnu.ac.kr}, \quad \footnotesize\texttt{hyungshin@cnu.ac.kr}
}

\renewcommand{\emph}[1]{\textit{#1}}
\begin{document}

\maketitle

\begin{abstract}
We propose Ciliary-DETR, a framework for test-time resolution adjustment analogous to biological accommodation.
While multi-scale data augmentation improves robustness to scale variation, modern detectors rely on fixed inference resolutions, potentially limiting flexibility and robustness.
Similar to the ciliary muscle, we introduce a lightweight scale predictor that dynamically estimates test-time scale factors across a wide range of input scales.
The core challenge is that the optimal input scale is inherently unobservable under standard training setups.
To address this challenge, we introduce a parametric formulation of desired scaling behavior, leading to loss-driven objectives that guide scale optimization.
Overall, our method enables flexible and efficient single-pass inference, bridging the gap between training-time robustness and test-time adaptation.




\end{abstract}

\section{Introduction}
\label{sec:intro}

Emulating the human visual system remains an important goal in the computer vision community~\cite{dosovitskiy2020image, carion2020end, gu2024mamba} due to its remarkable perceptual robustness. 
This robustness arises not only from accurate neural processing but also from the ability to flexibly regulate the sensory input itself (e.g., focal depth). 
However, such input-level adaptability is largely overlooked in modern detection networks~\citep{carion2020end, zhang2022dino, jia2023detrs, dong2025augdetr}, as they operate on a fixed input representation without sample-specific adaptability.
In particular, image resolution--one of the critical factors in visual representation--is typically treated as a static design choice rather than a dynamic, data-dependent variable.

\begin{figure*}[t]
    \centering
    \begin{subfigure}[b]{0.55\textwidth}
        \includegraphics[width=\linewidth]{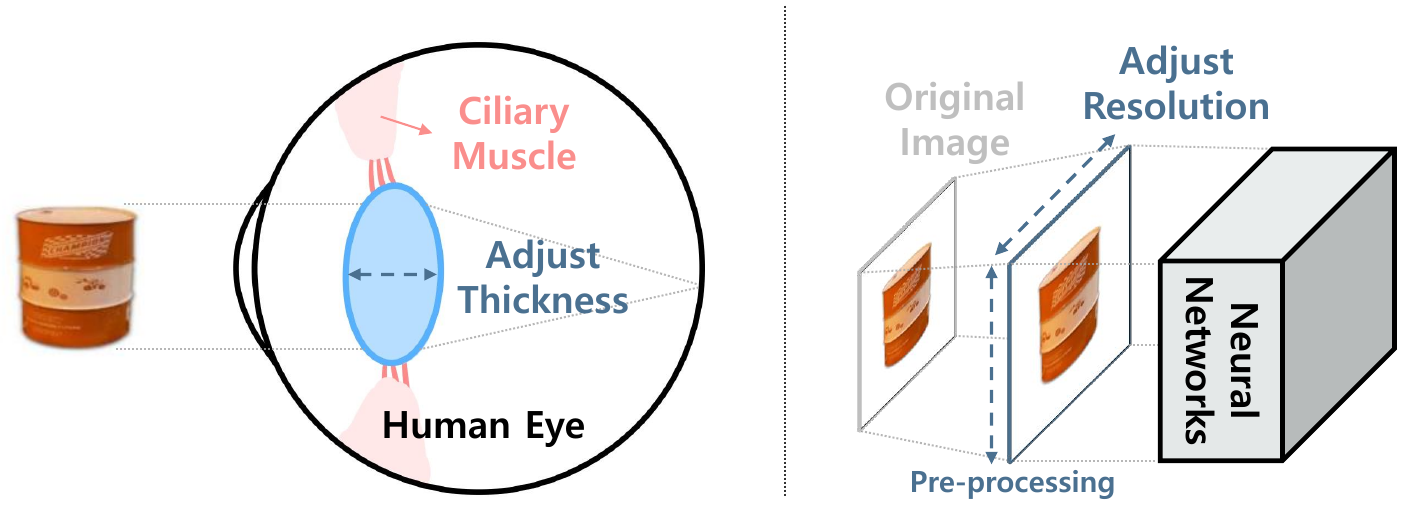}
        \caption{Conceptual Relationship}
        \label{fig:relation_to_lens}
    \end{subfigure}%
    \quad
    \begin{subfigure}[b]{0.4\textwidth}
        \includegraphics[width=\linewidth, trim={0cm 0.5cm 0cm 0cm}, clip]{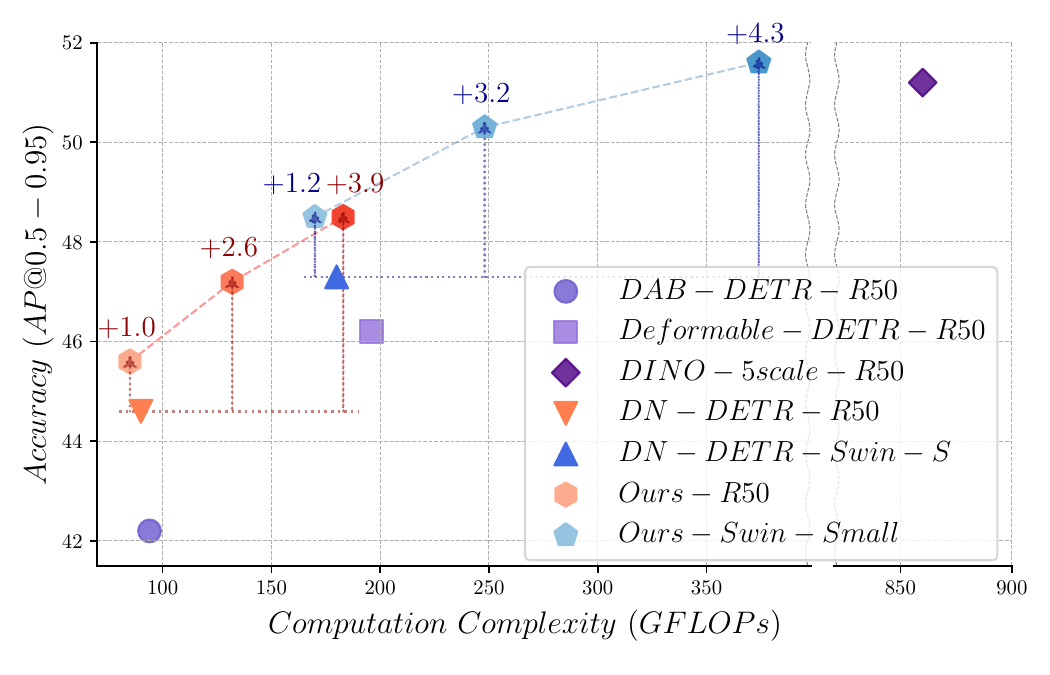}
        \caption{Performance Comparison}
        \label{fig:comp_others}
    \end{subfigure}
    \caption{
        (a) The change in the eye lens results in a modification of retinal projection size by regulating the focal depth.
        Similarly, this projection size is largely determined by image resolution in visual DNNs, leading to a functional analogy between biological mechanisms and DNN operations.
        (b) On MS COCO~\cite{lin2014microsoft}, our networks achieve consistent accuracy gains compared to DN-DETR baseline~\cite{li2022dn} with only image-level adaptation.
    }
\end{figure*}

In biological vision, a key mechanism for maintaining stable perception is visual accommodation~\cite{tortora2018principles}, which continuously adjusts the shape of the eye lens, resulting in changes in the scale of the retinal image.
As illustrated in \cref{fig:relation_to_lens}, we interpret image scale as an analogue of the eye lens by relating image resolution to scale and depth perception.
This leads to a simple but interesting perspective: we formulate image scale as a directly learnable variable that can be dynamically determined in a data-dependent manner.
This establishes a conceptual connection between biological vision and DNNs, leading to an efficient form of accommodation-like image-level scaling.
However, this introduces a key challenge: the optimal input scale is \textit{inherently unobservable} and lacks \textit{explicit supervision}, potentially making direct scale optimization ill-posed under standard training setups.

Previous methods for adaptive scaling often solved this problem by introducing architectural coupling, such as multi-branch designs~\citep{zhu2021dynamic} or multi-inference strategies~\citep{huang2017multi, yang2020resolution, wang2020glance, wang2021not, najibi2019autofocus}.
However, these approaches do not treat resolution as a learnable variable; instead, they rely on predefined resolutions.
Such methods often require different network structures for different resolution sets, which can hinder scalability.
Another line of work proposes a method that predicts an image-level histogram over scales~\citep{hao2017scale}.
This approach treats scale as a fixed-dimensional distributional parameter and requires a filtering step to obtain output scales.
Its scaling performance depends on the filtering strategy, which also conflicts with the end-to-end philosophy of modern detectors.

\begin{figure*}[t]
    \centering
    \includegraphics[width=\linewidth]{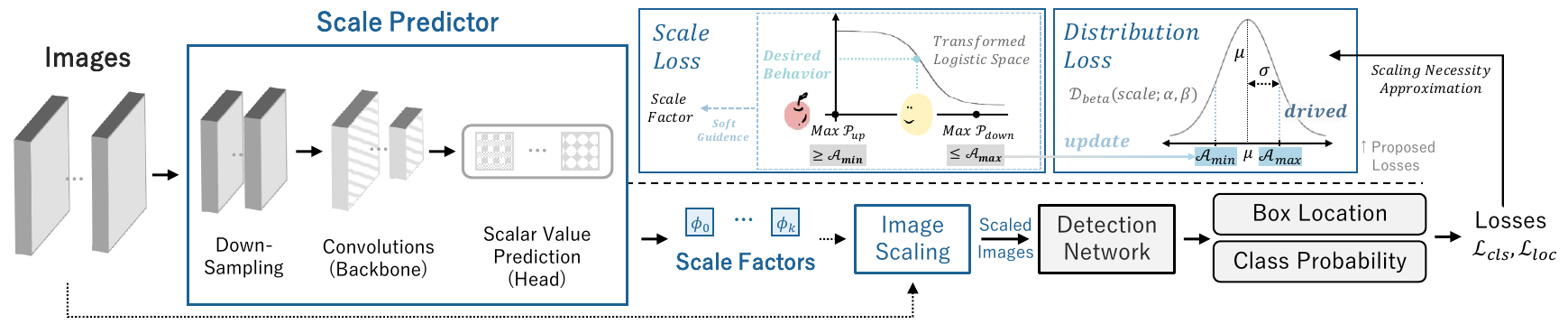}
    \caption{
        The overall architecture of Ciliary-DETR.
        A lightweight \textit{scale predictor} is placed prior to the detector to predict input-specific scale factors for test-time resolution adaptation. 
        To guide scale optimization, two complementary losses are introduced: \textit{scale loss} for size-aware adaptation and \textit{distribution loss} for performance-driven adaptation. 
    }
    \label{fig:overall_architecture}
\end{figure*}

In contrast, we propose \textit{Ciliary-DETR}, a framework that treats image resolution as a learnable variable within modern DETR-based detectors.
The image scales are flexibly adjusted by introducing a lightweight \textit{scale predictor}, directly estimating an input-dependent scale factor.
As shown in \cref{fig:overall_architecture}, this module enables scalable image processing prior to detection, analogous to a ciliary muscle that controls the eye lens.
This image adaptation is decoupled from task-specific prediction, thereby preserving the architectural simplicity of the detection pipeline.


The remaining challenge is how to learn adaptive and accurate scaling behavior without explicit supervision.
In this context, we introduce a parametric modeling of data-level scaling requirements, leading to probabilistic scale optimization.
Specifically, we provide two complementary objectives:
(1) a \textit{scale loss} that optimizes scale-dependent behavior through a probabilistic formulation, and 
(2) a \textit{distribution loss} that regularizes scaling bias by capturing the relationship between scale and detection performance.
Together, these objectives enable stable and data-driven scale adaptation without requiring ground-truth scaling annotations.

Instead of performing explicit multi-resolution inference, we aim to achieve efficient single-resolution optimization.
We enable multi-scale reasoning within a single forward pass by integrating multi-scale (MS) data augmentation~\cite{liu2016ssd,carion2020end}: (1) training the detector with MS augmentation and (2) jointly optimizing it with the scale predictor.
MS augmentation improves robustness to scale variation by exposing the model to diverse scale conditions,  enabling implicit multi-scale reasoning at inference time.
Overall, our formulation bridges training-time robustness and test-time adaptation, leveraging the modularity of our framework.


Unlike prior adaptive scaling methods, our framework directly controls the input scale, enabling test-time end-to-end scaling.
The proposed method allows straightforward deployment without requiring architectural modifications, which is effective for flexible input adaptation without hindering scalability.
Our scaling pipeline is inherently compatible with modern DETR-based detectors, in contrast to prior approaches built on conventional detectors~\citep{najibi2019autofocus, hao2017scale}.

The main contributions are summarized as follows:
(1) We propose a new formulation that treats image scale as a directly learnable variable.
(2) We introduce loss-driven optimization objectives that model data-driven scaling behavior without explicit supervision.
(3) We demonstrate that adaptive single-resolution inference can effectively approximate multi-scale reasoning, bridging training-time augmentation and test-time adaptation.
(4) We validate our framework through extensive experiments, demonstrating consistent gains in both accuracy and computational efficiency.


\section{Related Works}
\label{sec:related}

\noindent\textbf{Transformers for Object Detection.}
Traditional object detectors based on convolutional neural networks~\citep{lin2017focal, tian2019fcos, redmon2016you, bochkovskiy2020yolov4, wang2023yolov7} typically rely on hand-designed components such as anchor mechanisms~\citep{ren2015faster} or non-maximum suppression (NMS). 
DETR~\citep{carion2020end} introduced a transformer-based end-to-end paradigm that establishes direct optimization of predicted bounding boxes.
This motivates a series of follow-up works, including multi-scale feature aggregation~\citep{zhu2020deformable, zhang2023towards}, denoising-based training~\citep{li2022dn, zhang2022dino}, and query formulation strategies~\citep{meng2021conditional, liu2022dab, wang2022anchor}. 
More recently, dynamic DETR variants~\citep{roh2021sparse, dai2021dynamic, huang2024dq} have explored adaptive computation and representation within the network architecture. 
In contrast to these approaches, we investigate \textit{input-level adaptation} by directly optimizing the image resolution itself, providing an orthogonal direction to internal architectural adaptation.

\noindent\textbf{Optimizing Image Resolution in Object Detection.}
At training time, multi-scale data augmentation~\citep{liu2016ssd, beyer2023flexivit, dehghani2023patch, tian2023resformer} enables robust performance across scale variation and is widely adopted in dense prediction tasks~\citep{singh2018analysis, singh2018sniper, law2018cornernet, carion2020end}.
Complementary to this strategy, prior works have explored adaptive resolution for dynamic test-time inference, primarily in image classification~\citep{huang2017multi, zhu2021dynamic, yang2020resolution, wang2020glance, wang2021not}, while a few extensions have been explored for object detection within conventional frameworks.
Previous methods often introduced tight coupling with architectural design, such as multi-branch networks~\citep{zhu2021dynamic} or post-hoc scale selection via multi-resolution inference~\citep{huang2017multi, wang2020glance, wang2021not, najibi2019autofocus}.
They typically operate on pre-defined resolution choices, rather than directly learning data-dependent scaling policies.
Such approaches often lead to variations in architectural design across different input resolutions, thereby potentially limiting image scalability.
Similar to our work, a scale proposal network~\citep{hao2017scale} proposes input-level scale prediction by estimating a scale histogram.
While effective, it formulates the decision process as a filtering problem from the predicted histogram with NMS, conflicting with the end-to-end philosophy of DETR.
In contrast, our framework treats image resolution as a directly learnable variable and predicts an input-dependent scale factor. 
This enables flexible and efficient single-pass inference over a wide scale range, bridging the gap between training-time augmentation and test-time adaptation.

\section{Method}

\subsection{Predicting Adaptive Image Scale: Architecture of Scale Predictor}

\noindent\textbf{Overall Prediction Procedure.}
We decouple scale optimization from the detection framework by introducing a scale predictor that enables data-driven image scaling.
Given an input image $\mathcal{I}$, the scale predictor $\Omega$ estimates a scale factor $\phi = \Omega(\mathcal{I})$.
The image is then resized via bilinear interpolation and padded to maintain consistent batch dimensions.
The interpolated image is directly passed to the detector, enabling end-to-end image scaling.
We adopt DN-DETR~\citep{li2022dn} as the detection network to clearly isolate the effect of resolution adaptation, as it follows a single-scale architecture with stable and fast convergence.
Note that we aim to demonstrate the effectiveness of our scaling method, rather than pursuing state-of-the-art performance.

\noindent\textbf{Prediction Architecture of Scale Factor.}
The design philosophy of the scale predictor is simplicity and flexibility, and it is composed of two main components: 
(1) a backbone network for extracting visual characteristics from images, and (2) a prediction head for estimating a single scale factor.
The input image is internally down-sampled to reduce computational overhead and fed into a lightweight backbone network, ResNet-18~\cite{he2016deep}.
The output features from ResNet-18 are projected into a compact latent representation via a fully connected (FC) layer, followed by three transformer encoder blocks to capture global context for scale prediction.
A final FC layer produces a scalar output, followed by a sigmoid activation to constrain the range within [0, 1]. 

\begin{figure*}[t]
    \centering
    \begin{subtable}[b]{0.2\textwidth}
        \Large
        \aboverulesep=0ex 
        \belowrulesep=0ex 
        \renewcommand*\arraystretch{1.8}
        \resizebox{1\textwidth}{!}{
                \begin{tabular}{ l | c  c}
                    \toprule\rule{0pt}{1.1EM}
                    & MFLOPs & Params. \\
                    \midrule\rule{0pt}{1.1EM}
                    Backbone & 1.55 $\times 10^3$ & 0.68M \\
                    Head & 0.36 & 0.21M \\\hline
                    Total & 1.55 $\times 10^3$ & 0.89M \\
            \bottomrule
            \end{tabular}
        }
        \caption{Computational Complexity}
        \label{tab:complexity_break}
    \end{subtable}%
    \quad
    \begin{subfigure}[b]{0.5\textwidth}
        \includegraphics[width=\linewidth]{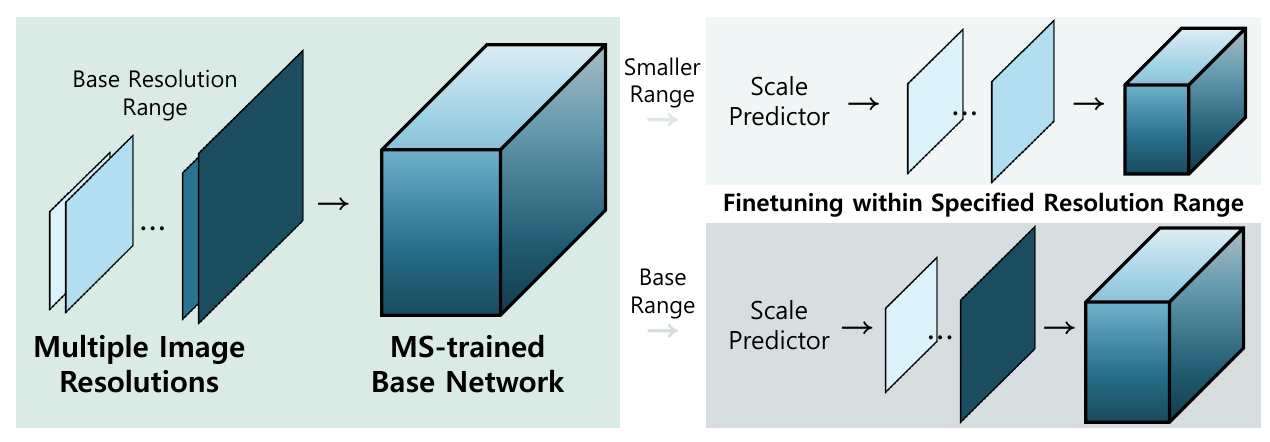}
        \caption{Training Strategy}
        \label{fig:trn_straetgy}
    \end{subfigure}%
    \quad
    \begin{subfigure}[b]{0.23\textwidth}
        \includegraphics[width=\linewidth, trim={0cm 0.2cm 0cm 0.8cm}, clip]{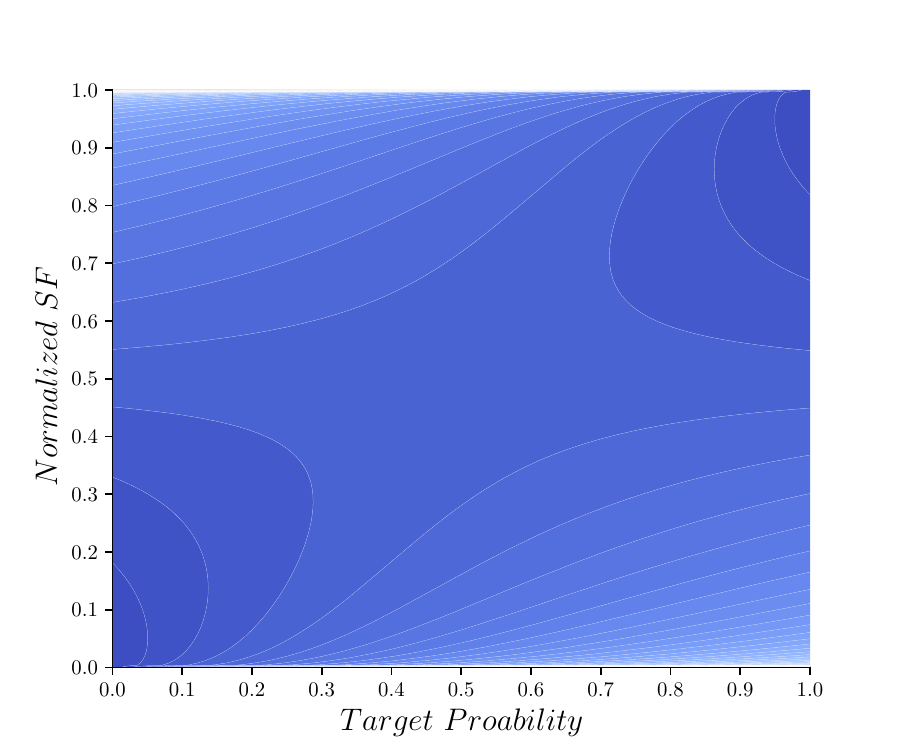}
        \caption{Loss Visualization}
        \label{fig:scale_loss}
    \end{subfigure}%
    \caption{
    (a) The obtained computational complexity of the scale predictor, which exhibits lower than 2 GFLOPs. 
    (b) Training strategy for multi-scale reasoning within a single resolution budget.
    Following the supernet philosophy, we adopt resolution variation between the two stages and use the same detection weights.
    (c) Visualization of \textit{scale loss}, exhibiting smooth and convex curvature.
    }
\end{figure*}

\noindent\textbf{Desired Range for Resolution Scaling.}
The output from the prediction head is mapped to a desired scaling range by specifying the minimum and maximum values, $\tau_{min}$ and $\tau_{max}$.
We transform the predicted value $\phi_{\sigma}$ to enforce the valid scaling range as $\phi = \max(\phi_{\sigma} \cdot \tau_{max},~\tau_{min})$.
The image resolution is then scaled according to $\phi$, followed by a discretization operation $f_{disc}$ that ensures the resolution is divisible by the granularity $\rho$.
The resulting resolution is computed as $\{\mathcal{W}',\mathcal{H}'\} = f_{disc} \big( \{ \mathcal{W},\mathcal{H} \} \cdot \phi, \rho \big)$.
As shown in \cref{tab:complexity_break}, the computational overhead of the scale prediction remains under 2 GFLOPs.
In practice, image resolution controls the network's trade-off between accuracy and computational cost.
Thus, adjusting $\tau_{min}$ and $\tau_{max}$ allows flexible control of the operating configuration without altering the detection pipeline.


\subsection{Training Adaptive Image Scale}
\label{ssec:trn_adaptive_res}

\noindent\textbf{Training Strategy.}
We operate under a single forward pass for both network flexibility and computational efficiency.
We promote multi-scale robustness under this setting by incorporating with multi-scale (MS) data augmentation.
This strategy exposes the network to randomly sampled scales, resulting in a improved robustness against scale variation.
As illustrated in \cref{fig:trn_straetgy}, we adopt a two-stage training strategy leveraging the modularity of our framework:
The detection network is initially trained with MS augmentation and then jointly updated with the scale predictor.

\noindent\textbf{Supernet-based Resolution Variation.}
From the perspective of neural architecture search~\citep{guo2020single, li2020gan, cai2019once}, the initial MS training shares similarities with supernet training, where multiple networks are extracted using different input resolutions (see \cref{ssec:more_resolution_info}).
Following this view, we adopt resolution variation and weight sharing during the joint training, similar to subnet training.
Specifically, MS training operates over a resolution range parameterized by $\tau_{min}$, $\tau_{max}$, and $\rho$. 
Next, the detector is jointly trained with a scale predictor under a sub-range $k$ defined by $\tau_{min}$, $\tau_{max}^k$, and $\rho$, where $\tau^k_{max} \leq \tau_{max}$. 
At this stage, all networks share the same initial supernet weights, leading to efficient training through parameter sharing.
The overall formulation enables efficient exploration of performance--efficiency trade-offs by simply adjusting the input resolution.

\noindent\textbf{Conceptual Analysis: Robustness under Single-pass Budget.}
The image-level scale variation can be roughly expressed as $\bar{s} + \delta_{s}$, where $\bar{s}$ and $\delta_s$ denote inter-image and intra-image scale variation, respectively.
Our scale loss provides robustness to inter-image variation $\bar{s}$ by optimizing a single-resolution scale, as discussed later.
Joint training further enables the detection network to handle intra-image variation $\delta_s$ by adapting its behavior to scaled inputs, leveraging the robustness acquired through MS augmentation.
Overall, this strategy enables implicit multi-scale reasoning under a single-resolution setting while avoiding the computational overhead of explicit multi-scale inference.



\subsection{Optimizing Adaptive Image Scale: Optimization Objectives for Input Scaling}
\noindent\textbf{Overview.}
We aim to define an instance-specific scaling objective to achieve normalized object scales by up-scaling small objects and down-scaling large ones.
However, object size alone cannot determine the appropriate scale factor; in particular, it is ambiguous which object sizes should correspond to a scale factor of 1.
To address this ambiguity, we formulate the scaling objective by viewing the scale predictor as a network that maps a probabilistic output to a positive scale factor.
Given this interpretation, we model data-level scaling preference in a probabilistic manner by introducing a soft saturation effect to stabilize scaling behavior:
\begin{equation}
\label{eq:global_rep}
    \mathcal{P}(s|\mathcal{A}_{min}, \mathcal{A}_{max}) \propto \sigma(s;\mathcal{A}_{min}, \mathcal{A}_{max}),
\end{equation}
where $s$ denotes the object size and $\mathcal{A}_{min}, \mathcal{A}_{max}$ are parameters that define the saturation region. 
This probabilistic interpretation enables soft alignment between object scales and scale factors, whose domains are inherently mismatched, e.g., normalized object scales $s \in (0,1]$ and scale factors $\phi \in [\tau_{min}, \tau_{max}]$.
To achieve adaptive data-driven scaling, we introduce two complementary objectives:
(i) a scale loss that defines the probabilistic optimization objective based on object scales, and
(ii) a distribution loss that aligns scaling decisions with downstream detection performance by updating the saturation parameters.
Our formulation directly encodes scaling objectives for reducing inter-image scale variation.
This allows the network to jointly optimize flexible input scaling and downstream prediction in a unified and stable manner.


\subsubsection{Scale Loss: Scale-specific Adaptation}
\label{ssec:scale_loss}

\noindent\textbf{Objectives for Scale Optimization.}
From a probabilistic perspective, data-level scaling preference can be modeled as a random variable that reflects how strongly an object requires scaling.
Specifically, this preference is modeled using a Bernoulli parameterization conditioned on the object size, with probabilities given by up-scaling $\mathcal{P}_{up}$ and its complement, down-scaling $\mathcal{P}_{down}$.
This formulation provides soft supervision that guides scale optimization, leading to an optimization objective based on \textit{binary cross-entropy (BCE)}.
This enables soft optimization behavior in ambiguous regions while maintaining a simple and stable objective.

\noindent\textbf{Probabilistic Scaling Formulation.}
Given the continuous nature of the scaling objective for achieving consistent object scales, we assume a monotonic relationship between object size and scaling preference.
Under the BCE-based formulation, we model this preference using a \textit{logistic parameterization} to capture its underlying probabilistic behavior.
This formulation encourages both predictions and supervision to follow a smooth sigmoid-shaped curve, enabling stable optimization.
The resulting sigmoid function represents a soft and monotonic preference within a bounded range.
The key challenge is how to incorporate this scaling preference into a concrete logistic parameterization.

\noindent\textbf{Introduction of Anchor Sizes.}
To this end, we introduce anchor sizes $\mathcal{A} = [\mathcal{A}_{min}, \mathcal{A}_{max}]$ that characterize the saturation effect of the scaling behavior: $\mathcal{P}_{up} \approx 1$ if $s \leq \mathcal{A}_{min}$ and $\mathcal{P}_{down} \approx 1$ if $s \geq \mathcal{A}_{max}$.
These conditions correspond to the saturation regions of the sigmoid function, allowing the logistic parameterization to be interpreted as a soft saturation effect governed by the anchor sizes.
To explicitly model this behavior, we define a domain transformation $\Lambda$ from the scale domain to the logit domain.

\noindent\textbf{Unified Probabilistic Formulation.}
Following the Bernoulli objective, scale optimization can be expressed using a single probabilistic variable by noting that $\mathcal{P}_{down} = 1 - \mathcal{P}_{up}$. 
We therefore focus on $\mathcal{P}_{up}$, which decreases monotonically with respect to the object size $s$.
For direct logistic representation under anchor sizes, $\Lambda$ is defined as an affine transformation:
\begin{equation}
    \Lambda(s) = s_n \cdot (s - \mathcal{A}_{min}) +\mathcal{X}_{min}, \quad
     s_n = \dfrac{\mathcal{X}_{max} - \mathcal{X}_{min}}{\mathcal{A}_{max} - \mathcal{A}_{min}},
\end{equation}
such that $\mathcal{A}_{min} \mapsto \mathcal{X}_{min}$ and $\mathcal{A}_{max} \mapsto \mathcal{X}_{max}$.
Here, we use a normalized object size $s\in (0, 1]$.
The up-scaling probability is then defined as $\mathcal{P}_{up} = \sigma \big( -\Lambda(s) \big)$, where the negative sign ensures a monotonically decreasing relationship.
The transformed range is determined using a small constant $\epsilon$ as $[\mathcal{X}_{min}, \mathcal{X}_{max}] = [\sigma^{-1}(\epsilon), \sigma^{-1}(1-\epsilon)]$, ensuring a smooth transition within a bounded range.
The anchor sizes are adaptively updated via the distribution loss described later.

\noindent\textbf{Scale Factor Optimization.}
The obtained probability expresses data-driven up-scaling preference over a continuous size domain.
For a unified probabilistic optimization framework, we normalize the scale factor as $\mathcal{P}_\phi = \phi / \tau_{max}$.
Here, $\phi$ is produced via a sigmoid-based transformation followed by scaling and clipping.
This normalization reinterprets the resulting scale factor as a probability-like quantity, enabling its use within a BCE objective in a manner analogous to a sigmoid output.
Given $\mathcal{P}_{down}=1-\mathcal{P}_{up}$, the scale loss balances the preference for up-scaling and down-scaling as follows:
\begin{equation} \label{eq:scale_loss}
\mathcal{L}_{scale}
= - \dfrac{1}{N} \sum_{i=1}^N
\mathbb{E}_{j\in \mathcal{I}_i} \left[
\mathcal{P}_{up}(s_j) \log \mathcal{P}_{\phi_i}
+
\mathcal{P}_{down}(s_j) \log (1 - \mathcal{P}_{\phi_i})
\right].
\end{equation}
In practice, a single scale factor $\phi_i$ is applied per image $\mathcal{I}_i$ and optimized over multiple instances $s_j \in \mathcal{I}_i$. 
This implicitly balances inter-instance variation by optimizing a single operating scale that reduces the average scale loss.
As illustrated in \cref{fig:scale_loss}, this loss exhibits a smooth and convex shape (see \cref{ssec:convex_scale_loss}) that reflects the inverse relationship between size and scaling.

\subsubsection{Distribution Loss: Performance-driven Alignment}
\label{ssec:dist_loss}

\noindent\textbf{Motivation.} 
To achieve accurate scale optimization, anchor sizes should be properly defined, but determining these without explicit scale supervision is non-trivial. 
To address this, we infer the network's scaling requirement directly from its detection performance, leveraging the strong dependency between performance and object scale. 
Inspired by the process of human eyeglass calibration, we estimate anchor sizes via three steps: 
(1) modeling the relationship between object scale and performance, 
(2) identifying a high-density region of the distribution, and 
(3) aligning this region with the network's global scaling capacity.
Specifically, we model the scale-performance trend as a parametric distribution conditioned on object sizes and fit this distribution to empirical performance observations within each image batch, leading to a distributional regression formulation.



\noindent\textbf{Parametric Distribution.} 
To model the scale-performance relationship, the parametric distribution should satisfy three key properties:
(1) it is defined over a bounded domain of normalized object sizes,
(2) it can represent asymmetric or skewed behaviors, and
(3) it allows stable parameterization with limited complexity.
To this end, we adopt a Beta distribution, $f_{beta}(s;\alpha, \beta)$, which naturally satisfies these requirements.
Unlike unbounded or symmetric alternatives, it provides the flexibility to model unimodal and skewed trends over the bounded interval $[0,1]$.
The parameters $\alpha$ and $\beta$ are treated as learnable variables and are optimized to capture the underlying scale-performance trend.

\noindent\textbf{Observation Distribution.}
To obtain empirical observations of the scale-performance relationship, we construct a loss-driven distribution. 
Given the up-scaling preference that decreases monotonically with object size, we use detection loss as a coarse proxy for inverse detection quality, where detection performance generally improves with increasing object size~\cite{carion2020end, meng2021conditional, zhang2022dino}.
We define a discrete observation distribution over object sizes by normalizing detection losses within an image batch $b$:
\begin{equation}
 \mathcal{D}_{obs}(s_i) = {\mathcal{L}'_{det}(s_i)}/{\sum\nolimits_{j\in b} \mathcal{L}'_{det}(s_j)},   
\end{equation}
where $\mathcal{L}'_{det}$ denotes the detection loss detached from gradient propagation.
This reflects the relative performance degradation across object scales, which can be interpreted as a discrete probability mass distribution.

\noindent\textbf{Distributional Regression.} 
The parametric distribution is then updated by minimizing its discrepancy from the observation distribution.
For geometry-aware optimization over the size domain $s$, we employ a 1-Wasserstein distance~\citep{lv2024wasserstein, ramdas2017wasserstein} to capture the discrepancy between the two distributions.
The distribution loss is formulated as,
\begin{equation} 
\label{eq:dist_loss} 
\mathcal{L}_{dist}(\alpha,\beta) = W_{1} (\mathcal{D}_{beta}, \mathcal{D}_{obs}),
\end{equation} 
where $\mathcal{D}_{beta}$ is a discrete distribution obtained by evaluating $f_{beta}(s;\alpha, \beta)$ on object sizes $s_i \in b$ and normalizing the resulting values.
This discrete approximation ensures that both distributions share the same support, enabling stable optimization.
As depicted in \cref{fig:dist_plot}, the learned parametric distribution closely matches the loss-driven observation distribution. 


\noindent\textbf{Estimating Anchor Sizes.}
After fitting the distribution, we estimate the high-density region using a practical approximation given by $[\mathcal{A}_{min}, \mathcal{A}_{max}]=[\mu - \sigma, \mu + \sigma]$, where $\mu$ and $\sigma$ denote the mean and standard deviation, respectively.
This approximation provides a simple yet effective characterization of the dominant region of the distribution, as discussed in \cref{ssec: beta_distribution}.

\noindent\textbf{Alignment with Target Capacity.}
A limitation of the loss-driven representation is that it lacks global consistency across different network configurations, making the resulting scale optimization dependent on observed performance alone. 
To address this, we incorporate the network's scaling capacity following our supernet-based resolution framework. 
Specifically, we adjust the estimated anchor range by rescaling its lower bound as $[\xi \cdot \mathcal{A}_{min}, \mathcal{A}_{max}]$,
where $\xi = \tau_{max} / \tau_{max}^k$ denotes the relative scaling capacity of network $k$.
This adjustment aligns the inferred scale range with the supernet capacity, enabling more aggressive up-scaling for lower-capacity networks while preserving consistency across configurations.





\begin{table*}[t]
    \centering
    \Large
    \renewcommand*\arraystretch{1.6}
    \aboverulesep=0ex 
    \belowrulesep=0ex 
    \caption{
        Comparison on MS COCO2017 \texttt{val}~\citep{lin2014microsoft}. 
        $\circ$ and ${\diamond}$ denote our re-implemented results and the use of our subnet-like strategy, respectively.
        In \# epochs, ${\dagger}$ and ${\ddagger}$ denote first (MS aug.) and second (joint) training phases.
        GFLOPs are averaged across the validation set with end-to-end measurements.
    }
    \label{tab:acc}
    \resizebox{\linewidth}{!}{
        \begin{tabular}{l | c c | c c c c c c | c c | c c}
            \toprule\rule{0pt}{1.1EM}
             & Strategy$_{img}$ & Resolution & AP & AP$_{50}$ & AP$_{75}$ & AP$_s$ & AP$_m$ & AP$_l$ & GFLOPs & \# Params & \multicolumn{2}{c}{\# Epochs} \\
            \midrule\rule{0pt}{1.1EM}
            DN-DETR-R50~\citep{li2022dn} & MS training &  & 44.1 & 64.4 & 46.7 & 22.9 & 48.0 & 63.4 & 94 & 44M & \multicolumn{2}{c}{50} \\
            DN-DETR-R50$^\circ$ & MS training & & 44.6 & 64.8 & 47.7 & 23.9 & 48.5 & 63.3 & 90 & 44M & \multicolumn{2}{c}{50} \\
            \rowcolor{UltraLightGray} & & \cellcolor{White} & 44.5 & 64.7 & 47.2 & 23.4 & 48.2 & 64.2 & \textbf{75} & 45M & \cellcolor{White} & 10$^\ddagger$ \\ 
            \rowcolor{UltraLightGray}\multirow{-2}{*}{\texttt{+ Ciliary-DETR}} & \multirow{-2}{*}{Ours ($\tau_{max}=1.35$)} & \cellcolor{White}\multirow{-4}{*}{[480, 800]} & 44.6 & 65.0 & 47.3 & 23.0 & 48.2 & 65.0 & \textbf{74} & 45M & \cellcolor{White}\multirow{-2}{*}{40$^\dagger$ + } & 20$^\ddagger$\\
            \hline\hline
            DN-DETR-Wide-R50$^{\circ}$ & MS training & & 47.2 & 66.9 & 51.1 & 28.8 & 50.8 & 61.8 & 226 & 44M & \multicolumn{2}{c}{50} \\
            \rowcolor{DarkGray} & & \cellcolor{White} & \textbf{48.2} & 67.9 & 52.0 & 29.0 & 51.8 & 65.1 & \textbf{182} & 45M & \cellcolor{White} & 10$^{\ddagger}$\\ 
            \rowcolor{DarkGray} & & \cellcolor{White} & \textbf{48.2} & 68.0 & 52.3 & 28.9 & 52.1 & 64.2 & \textbf{186} & 45M & \cellcolor{White} & 20$^{\ddagger}$\\ 
            \rowcolor{DarkGray}\multirow{-3}{*}{\texttt{+ Ciliary-DETR-Base}} & \multirow{-3}{*}{Ours ($\tau_{max}=2.15$)} & \cellcolor{White}\multirow{-4}{*}{[320, 1280]} & \textbf{48.5} & 68.5 & 52.3 & 29.1 & 52.5 & 65.5 & \textbf{183} & 45M & \cellcolor{White} & 32$^{\ddagger}$ \\
            \cline{3-3}
            \rowcolor{LightGray} & & \cellcolor{White} & \textbf{47.2} & 67.4 & 50.9 & 28.3 & 50.9 & 64.4 & \textbf{132} & 45M & \cellcolor{White} & 10$^{\ddagger}$ \\
            \rowcolor{LightGray}\multirow{-2}{*}{\texttt{+ Ciliary-DETR-Medium}$^{\diamond}$} & \multirow{-2}{*}{Ours ($\tau_{max}=1.80$)} & \cellcolor{White}\multirow{-2}{*}{[320, 1088]} & \textbf{47.2} & 67.2 & 50.8 & 27.3 & 50.8 & 64.8 & \textbf{132} & 45M & \cellcolor{White} & 20$^{\ddagger}$ \\
            \cline{3-3}
            \rowcolor{VeryLightGray} & & \cellcolor{White} & \textbf{45.4} & 65.8 & 48.3 & 25.6 & 49.2 & 64.6 & \textbf{86} & 45M & \cellcolor{White} & 10$^{\ddagger}$ \\
            \rowcolor{VeryLightGray}\multirow{-2}{*}{\texttt{+ Ciliary-DETR-Small}$^{\diamond}$} & \multirow{-2}{*}{Ours ($\tau_{max}=1.45$)} & \cellcolor{White}\multirow{-2}{*}{[320, 864]} & \textbf{45.6} & 66.1 & 48.5 & 25.6 & 49.9 & 64.6 & \textbf{85} & 45M & \cellcolor{White}\multirow{-7}{*}{40$^{\dagger}$ + } & 20$^{\ddagger}$ \\
            \hline \hline
            DN-DETR-Swin-Small$^\circ$ & MS training & & 47.3 & 68.5 & 50.3 & 25.9 & 51.4 & 67.8 & 180 & 69M & \multicolumn{2}{c}{50} \\
            \rowcolor{UltraLightGray} {\texttt{+ Ciliary-DETR}} & Ours ($\tau_{max}=1.35$) & \cellcolor{White}\multirow{-2}{*}{[480, 800]} & \textbf{47.4} & 69.7 & 50.8 & 27.2 & 51.6 & 68.0 & \textbf{148} & 69M & 40$^\dagger$ + & 20$^\ddagger$\\
            \hline\hline
            DN-DETR-Wide-Swin-S$^\circ$ & MS training & & 50.5 & 70.6 & 54.6 & 32.4 & 54.6 & 67.2 & 450 & 44M & \multicolumn{2}{c}{50} \\
            \rowcolor{DarkGray}\multirow{-1}{*}{\texttt{+ Ciliary-DETR-Base}} & \multirow{-1}{*}{Ours ($\tau_{max}=2.15$)} & \cellcolor{White}\multirow{-2}{*}{[320, 1280]} & \textbf{51.6} & 71.6 & 56.0 & 32.1 & 56.2 & 69.4 & \textbf{374} & 70M & \cellcolor{White} & 32$^{\ddagger}$ \\
            \rowcolor{LightGray} \texttt{+ Ciliary-DETR-Medium}$^{\diamond}$ & Ours ($\tau_{max}=1.80$) & [320, 1088] & \textbf{50.3} & 70.8 & 54.4 & 30.9 & 54.6 & 69.0 & \textbf{248} & 70M & \cellcolor{White} & 20$^{\ddagger}$ \\
            \rowcolor{VeryLightGray} \texttt{+ Ciliary-DETR-Small}$^{\diamond}$ & Ours ($\tau_{max}=1.45$) & [320, 864] & \textbf{48.5} & 69.1 & 51.9 & 28.3 & 53.0 & 68.1 & \textbf{170} & 70M & \cellcolor{White}\multirow{-3}{*}{40$^{\dagger}$ + } & 20$^{\ddagger}$\\
            \bottomrule
        \end{tabular}
    }
\end{table*}

\begin{table}[t]
    \begin{minipage}[t][][b]{0.48\linewidth}
        \vspace*{0mm}
        \centering
        \includegraphics[width=\linewidth]{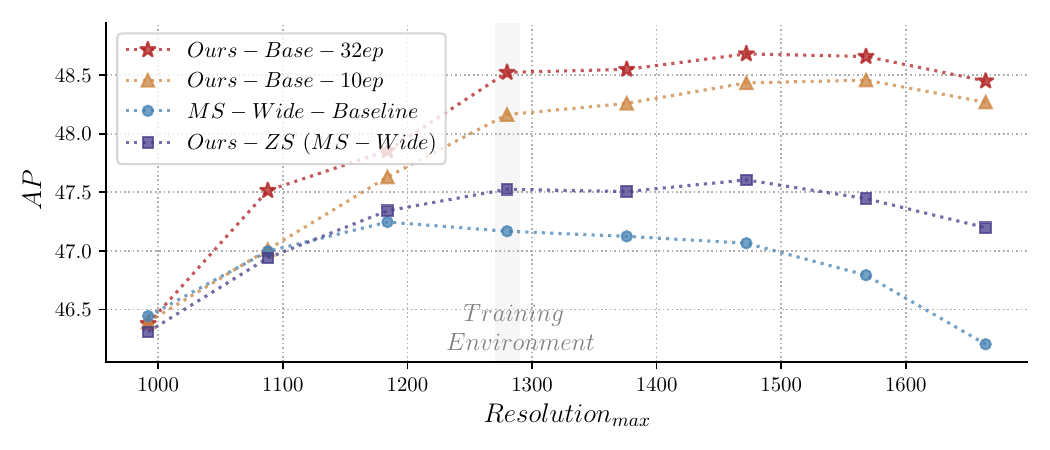}
        \captionof{figure}{
            Test-time resolution variation using networks (R50).
            Ours-ZS: Zero-shot use of the scale predictor on MS-Wide baseline.
        }
        \label{fig:test-time_res_variation}
        \vspace*{0mm}
    \end{minipage}
    \hfill
    \begin{minipage}[t][][b]{0.49\linewidth} 
        \vspace*{0mm}
        \centering
        \Large
        \renewcommand*\arraystretch{1.6}
        \aboverulesep=0ex 
        \belowrulesep=0ex 
        \caption{  
            Zero-shot use of the scale predictor (Ciliary-DETR-R50) on trained detectors.
        }
        \label{tab:zeroshot_other_detectors}
        \resizebox{\linewidth}{!}{
            \begin{tabular}{l | c | c c || c | c c}
            \toprule\rule{0pt}{1.1EM}%
             & Epochs & AP & GFLOPs & Epochs & AP & GFLOPs \\
            \midrule\rule{0pt}{1.1EM}%
            Deform-DETR-R50~\citep{zhu2020deformable} & 50 & \textbf{47.0} & 172 & - & - & - \\
            \texttt{Token Sparsity (50\%)}~\citep{roh2021sparse} & 50 & 46.3 & \textbf{136} & - & - & - \\
            \rowcolor{UltraLightGray} \texttt{+ Ciliary-DETR-Zeroshot} & 50 & \textbf{47.0} & \textbf{141} & - & - & - \\\hline
            DINO-DETR-4sc-R50~\cite{zhang2022dino} & 12 & 49.2 & 246 & 24 & \textbf{50.6} & 246 \\
            \texttt{Dynamic \#Queries (5sc)}~\citep{huang2024dq} & - & - & - & 24 & 50.2 & - \\
            \rowcolor{UltraLightGray} \texttt{+ Ciliary-DETR-Zeroshot} & 12 & \textbf{49.4} & \textbf{202} & 24 & 50.5 & \textbf{202} \\\hline
            $\mathcal{H}$-DETR-R50~\cite{jia2023detrs} & 12 & \textbf{49.1} & 236 & 36 & \textbf{50.3} & 236 \\    
            \rowcolor{UltraLightGray} \texttt{+ Ciliary-DETR-Zeroshot}  & 12 & \textbf{49.1} & \textbf{192} & 36 & \textbf{50.3} & \textbf{192} \\
            \bottomrule
        \end{tabular}
        }
        \vspace*{0mm}
    \end{minipage}
\end{table}

\section{Experiments}

\subsection{Implementation Details} 
\label{ssec:impl_detail}
We use a multi-scale (MS) training baseline with input resolutions ranging from 480 to 800 with a step size of 32, resulting in 11 resolution candidates.
For our method, we adopt a wider range of 320 to 1280 with the same step size, yielding 31 candidates.
A narrower sub-range is applied within this wide range, with maximum resolutions of 864 and 1088 for small and medium networks, respectively.
We train the model for 40 epochs during the initial MS training stage, while the joint training stage uses 10--32 epochs to study the convergence behavior of the scale predictor.
More implementation details are provided in \cref{sec:more_impl_details}.


\subsection{Main Results}
\noindent\textbf{Detection Performance.}
\cref{tab:acc} shows strong performance-efficiency trade-offs for our method.
Under the [480, 800] setting, our method reduces GFLOPs by \textit{17\%} compared to the baseline while maintaining comparable AP.
Notably, Ciliary-DETR-Base (i.e., ours-B) achieves \textit{+1.3/+1.1 AP} with \textit{19\%/17\% fewer GFLOPs} compared to the R50/Swin-S wide baselines, respectively.
As shown in \cref{fig:comp_others}, these models outperform other strong baselines.
This setup improves AP across all object scales, yielding notable gains for large objects without degrading performance on small objects.
These results suggest that our scale predictor effectively learns adaptive scale factors across a wide resolution range.
Moreover, ours-B-R50 achieves comparable AP with only 10 epochs relative to 20 epochs, indicating fast convergence.
Our supernet-like training further improves the performance--efficiency trade-off: ours-Small outperforms the [480, 800] baseline by at least \textit{+1.0 AP} while using fewer GFLOPs.
Furthermore, \cref{tab:lvis_result} shows substantial improvements on LVIS with \textit{+3.5/+1.4 AP}.

\begin{table*}[t]
    \centering
    \caption{
        (b) $\Delta$ indicates the difference from the original adaptation.
        (c) $\star$ denotes test-time adjustment.
    }
    \begin{subtable}[t]{0.58\textwidth}
        \renewcommand*\arraystretch{1.6}
        \aboverulesep=0ex 
        \belowrulesep=0ex 
        \caption{Performance comparison on \texttt{LVISv1.0}~\cite{gupta2019lvis}.}
        \label{tab:lvis_result}
        \resizebox{\linewidth}{!}{
            \begin{tabular}{l | c | c c c c c c}
            \toprule\rule{0pt}{1.1EM}
             & Resolution & AP & AP$_{50}$ & AP$_{75}$ & AP$_{s}$ & AP$_{m}$ & AP$_{l}$ \\
            \midrule\rule{0pt}{1.1EM}
            DN-DETR-R50 & [480, 800] & 21.5 & 32.5 & 22.4 & 13.0 & 29.2 & 39.4 \\
            DN-DETR-Wide-R50 & [320, 1280] & 22.6 & 33.4 & 24.0 & 15.6 & 30.0 & 36.9 \\
            \rowcolor{DarkGray} \texttt{+ Ciliary-DETR-Base-20ep} & [320, 1280] & \textbf{24.0} & 34.6 & 25.5 & 16.0 & 32.1 & 40.5 \\
            \bottomrule
            \end{tabular}
        }
        \begin{subtable}[t]{\textwidth}
            \vspace{0.1cm}
            \Huge
            \begin{subtable}[t]{0.53\textwidth}
                \caption{
                    Our networks (R50) tested on fixed/max resolution.
                }
                \label{tab:max_inf}
                \renewcommand*\arraystretch{1.6}
                \aboverulesep=0ex 
                \belowrulesep=0ex 
                \resizebox{\linewidth}{!}{
                    \begin{tabular}{l c c}
                        \toprule\rule{0pt}{1.1EM}
                         & AP$_{\color{gray}\mathbf{\Delta_{AP}}}$ & GFLOPs$_{\color{gray}\mathbf{\Delta_{GFLOPs}}}$ \\
                        \midrule\rule{0pt}{1.1EM}
                        Fixed-Ours-B-10ep & 47.2$_{\color{Maroon}\textbf{-1.0}}$ & 226$_{\color{Maroon}\textbf{1.2}\times}$ \\
                        Fixed-Ours-B-32ep & 47.9$_{\color{Maroon}\textbf{-0.6}}$ & 226$_{\color{Maroon}\textbf{1.2}\times}$ \\
                        \bottomrule
                    \end{tabular}
                }
            \end{subtable}%
            \hfill
            \begin{subtable}[t]{0.45\linewidth}
                \caption{Matching accuracy (R50).}
                \label{tab:match_acc}
                \renewcommand*\arraystretch{1.6}
                \aboverulesep=0ex 
                \belowrulesep=0ex 
                \resizebox{\linewidth}{!}{
                    \begin{tabular}{l c c c}
                        \toprule\rule{0pt}{1.1EM}
                         & Res. & AP & GFLOPs \\
                        \hline
                        MS-Wide-Baseline & 1280 & 47.2 & 226 \\
                        \rowcolor{DarkGray} \texttt{Ours-Base-10ep} & 1136$^{\star}$ & 47.2 & \textbf{142} \\ 
                        \rowcolor{DarkGray} \texttt{Ours-Base-32ep} & 1076$^{\star}$ & 47.3 & \textbf{129} \\
                        \bottomrule
                    \end{tabular}
                }
            \end{subtable}
        \end{subtable}
    \end{subtable}%
    \hfill
    \begin{subtable}[t]{0.4\textwidth}
        \begin{subtable}[t]{\textwidth}
            \large
            \renewcommand*\arraystretch{1.55}
            \aboverulesep=0ex 
            \belowrulesep=0ex 
            \caption{
                Comparison with test-time multi-scale strategies with a 96 resolution granularity.
                We select the best resolution for BFS (brute-force search) and merge predictions for MS TTA (multi-scale test-time augmentation).
            }
            \label{tab:comp_with_mstrn}
            \resizebox{\linewidth}{!}{
                \begin{tabular}{l | c | c || c}
                \toprule\rule{0pt}{1.1EM}
                 & Resolution & AP (R50) & AP (Swin-S) \\
                \hline
                Baseline-Wide & 1280 & 47.2$_{\color{white}\textbf{-0.0}}$ & 50.5$_{\color{white}\textbf{-0.0}}$ \\
                + \texttt{Random} & [800, 1280] & 45.6$_{\color{Maroon}\textbf{-1.6}}$ & 49.5$_{\color{Maroon}\textbf{-1.0}}$ \\
                + \texttt{MS TTA} & [800, 2336] & 47.5$_{\color{MidnightBlue}\textbf{+0.3}}$ & 50.8$_{\color{MidnightBlue}\textbf{+0.3}}$ \\
                + \texttt{BFS (IoU$_\text{Hungarian}$)} & [320, 1280] & 49.0$_{\color{MidnightBlue}\textbf{+1.8}}$ & 52.0$_{\color{MidnightBlue}\textbf{+1.5}}$ \\
                \rowcolor{DarkGray} \texttt{Ciliary-DETR-Base} & [320, 1280] & \textbf{48.5}$_{\color{MidnightBlue}\textbf{+1.3}}$ & \textbf{51.6}$_{\color{MidnightBlue}\textbf{+1.1}}$ \\
                \bottomrule
            \end{tabular}
            }
        \end{subtable}%
        \vspace{0.1cm}
    \end{subtable}
\end{table*}

    
        
    
    

\noindent\textbf{Test-time Generalization.}
\cref{fig:test-time_res_variation} demonstrates strong test-time generalization across a wide range of resolutions.
Our method consistently outperforms the wide baseline, particularly at high resolutions, achieving \textit{+2.2/+2.1 AP} at a resolution of 1664 for models trained for 10 and 32 epochs, respectively.
Applying zero-shot dynamic scaling to the baseline also improves robustness, yielding \textit{+1.0 AP} at 1664.
As shown in \cref{tab:zeroshot_other_detectors}, applying this strategy to unseen networks reduces GFLOPs by \textit{18–19\%} while maintaining AP.
Our method achieves performance comparable to token sparsification and higher AP than dynamic query methods, highlighting the benefit of data-driven scaling.

\noindent\textbf{Comparison with Test-time MS Strategies.}
\cref{tab:comp_with_mstrn} compares our method with existing test-time multi-scale strategies.
Random selection does not generalize well at inference time despite its effectiveness during training.
MS TTA (i.e., explicit multi-scale inference) shows limited effectiveness of naive multi-scale usage, even at much higher resolutions, as detailed in \cref{tab:ms_tta_impl}.
BFS (i.e., optimal resolution selection using ground-truth matching~\citep{carion2020end,kuhn1955hungarian}) achieves notable gains of +1.8/+1.5 AP within a similar resolution range to ours.
These results demonstrate the potential of single-resolution selection and serve as an approximate upper bound for optimization performance.
Notably, our method achieves results close to BFS, with only a 0.4–0.5 AP gap, suggesting that our method effectively approximates optimal resolution selection in a fully learnable manner.

\subsection{Analysis}

\noindent\textbf{Controlled Experiments.}
\cref{tab:max_inf} highlights the test-time benefits of our method in terms of both AP and GFLOPs.
Without dynamic scaling, our network incurs a drop of \textit{-1.0/-0.6 AP} while requiring approximately \textit{1.2$\times$ more GFLOPs}. 
This indicates that the performance gains stem not only from variable-resolution training but also from test-time adaptation.
\cref{tab:match_acc} demonstrates a comparison under a similar AP budget.
Our networks achieve substantially lower GFLOPs than the baseline, even when trained for only 10 epochs, resulting in a \textit{37\%} reduction in computation.

\noindent\textbf{Class-wise Comparison.}
\cref{tab:cls_comp} provides a class-wise comparison between ours-Base-R50 and the MS-Wide baseline, demonstrating strong generalization across diverse object categories.
Our method outperforms the baseline on 71 classes, with a maximum gain of +6.7 AP, while only 9 classes show performance degradation, with the largest drop of -2.5 AP.
As illustrated in \cref{fig:img_examples}, adaptive scaling is particularly beneficial for visually complex scenes, whereas simpler scenes may show marginal performance drops.
These results highlight the effectiveness of our single-pass multi-scale reasoning framework for handling diverse scenarios without architectural modifications.


\begin{table}[t]
    \begin{minipage}[b][][b]{0.4\linewidth}
        \vspace*{0mm}
        \centering
        \Huge
        \caption{Class-wise comparison.}
        \label{tab:cls_comp}
        \renewcommand*\arraystretch{1.6}
        \aboverulesep=0ex 
        \belowrulesep=0ex 
        \resizebox{\linewidth}{!}{
            \begin{tabular}{l | c | c}
                \toprule\rule{0pt}{1.1EM}
                 & \# & Top-3 Classes $_{\color{gray}\mathbf{\Delta}\textbf{AP}}$ \\
                \midrule\rule{0pt}{1.1EM}
                Positive Classes & 71 & toaster$_{\color{MidnightBlue}\textbf{+6.7}}$, parking meter$_{\color{MidnightBlue}\textbf{+5.1}}$, teddy bear$_{\color{MidnightBlue}\textbf{+4.0}}$ \\
                Negative Classes & 9 & bear$_{\color{Maroon}\textbf{-2.5}}$, donut$_{\color{Maroon}\textbf{-1.7}}$, vase$_{\color{Maroon}\textbf{-1.2}}$ \\
                \bottomrule
            \end{tabular}
        }
        \vspace*{0mm}
    \end{minipage}
    \hfill
    \begin{minipage}[b][][b]{0.58\linewidth} 
        \vspace*{0mm}
        \centering
        \includegraphics[width=\linewidth]{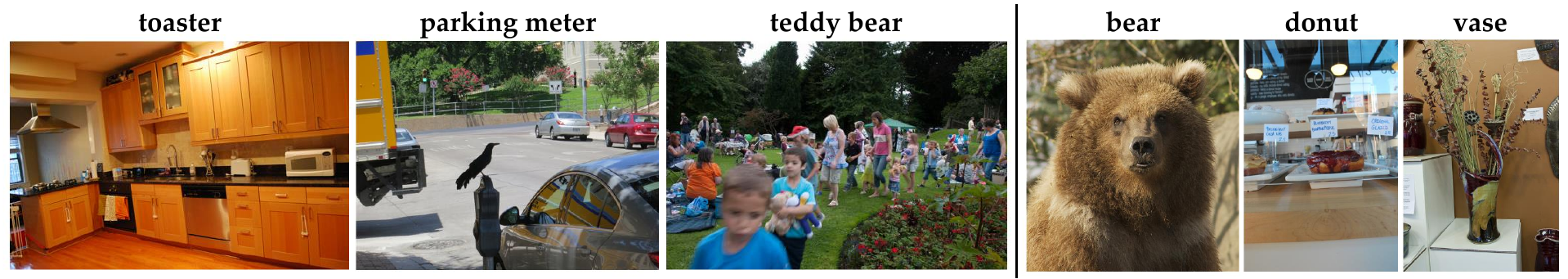}
        \captionof{figure}{
            (Left) Positive / (Right) Negative class examples.
        }
        \label{fig:img_examples}
        \vspace*{0mm}
    \end{minipage}
\end{table}

\begin{figure*}[t]
    \centering
    \begin{subfigure}[b]{0.30\textwidth}
        \includegraphics[width=\linewidth]{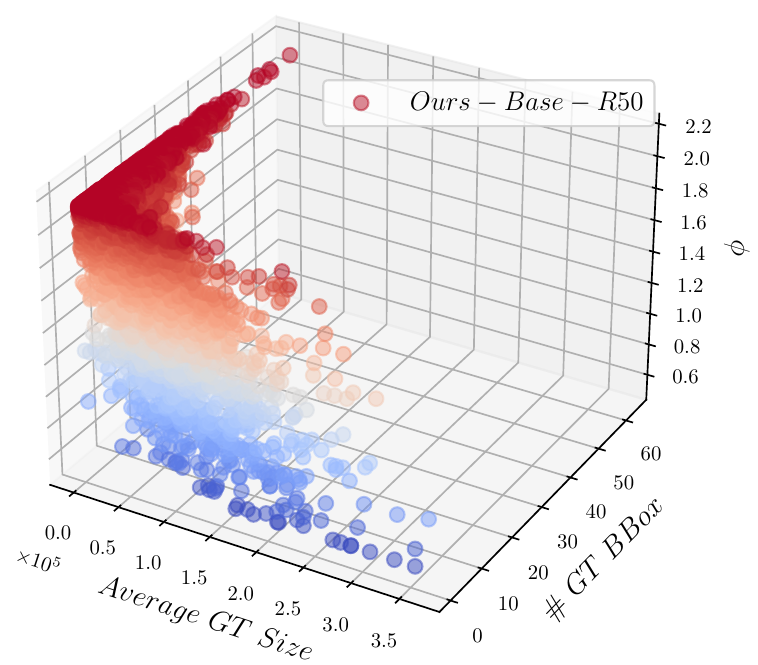}
        \caption{Predicted Scale Factors}
        \label{fig:sf_plot}
    \end{subfigure}%
    \quad
    \begin{subfigure}[b]{0.27\textwidth}
        \includegraphics[width=\linewidth, trim={0cm 0cm 0cm 0cm}, clip]{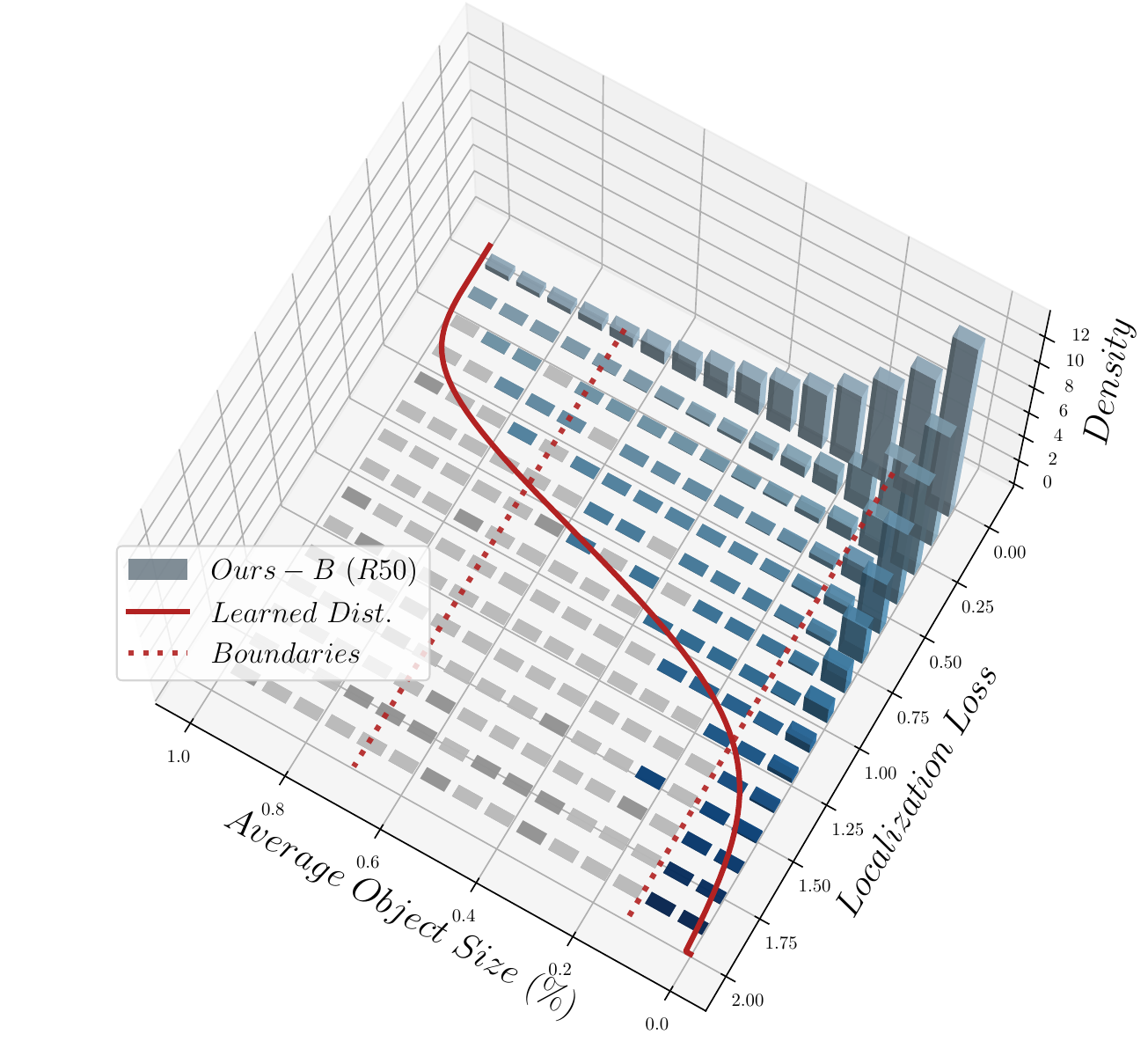}
        \caption{Obtained Scaling Anchors}
        \label{fig:dist_plot}
    \end{subfigure}%
    \quad
    \begin{subfigure}[b]{0.37\textwidth}
        \includegraphics[width=\linewidth]{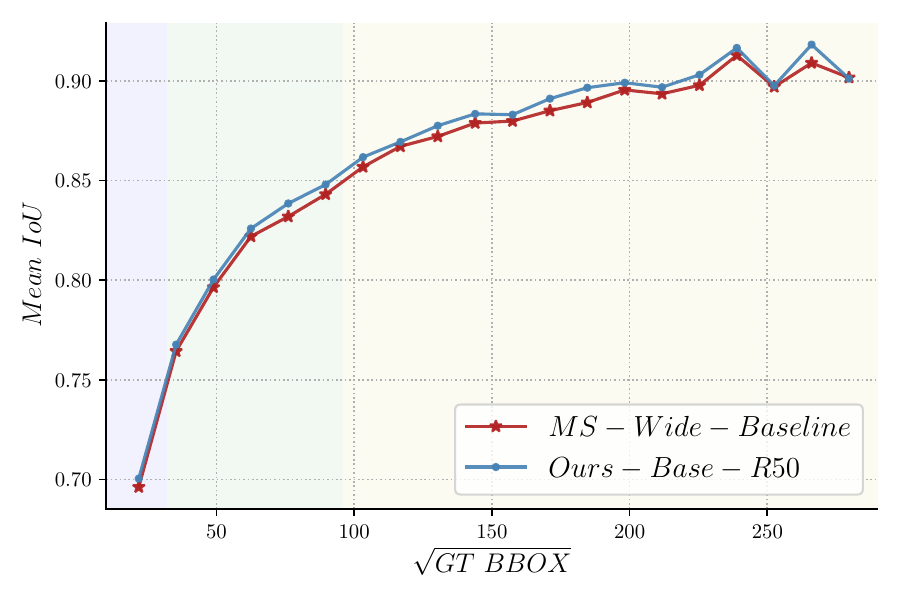}
        \caption{IoU Comparison}
        \label{fig:iou_comp_baseline}
    \end{subfigure}
    \caption{
    (a) Predicted scale factors with respect to object size and the number of objects in the image.
    (b) Visualization of the learned Beta distribution and anchor sizes with the loss distribution. 
    (c) Per-scale IoU comparison across various scales using Hungarian matching between ground truth and predictions.
        The background color indicates the size ranges of the COCO metrics.
    }
\end{figure*}

\noindent\textbf{Predicted Scale Factor.}
\cref{fig:sf_plot} shows that our scale predictor effectively estimates a wide range of scale factors.
The scale factor is inversely related to object size, assigning larger values to smaller objects.
It also tends to increase with the number of instances, indicating that the predictor adapts its behavior to different object distributions while capturing inter-image variation.

\noindent\textbf{Distribution Learning.}
\cref{fig:dist_plot} shows that the learned Beta distribution closely aligns with the loss-driven distribution.
This alignment enables the model to identify performance-sensitive size regions, from which effective anchor sizes are derived.

\noindent\textbf{IoU Comparison.}
As shown in \cref{fig:iou_comp_baseline}, our method consistently outperforms the wide baseline across various scales, indicating strong robustness to scale variation.
The adaptive scaling induced by the learned anchors avoids performance degradation for specific object sizes, demonstrating the effectiveness of the proposed method.

\subsubsection{Ablations}

\begin{table}[ht!]
    \vspace{-16pt}
    \begin{minipage}[b][][b]{0.33\linewidth} 
    \vspace*{0mm}
    \renewcommand*\arraystretch{1.5}
        \caption{Proposed Losses}
        \label{tab:eff_losses}
        \aboverulesep=0ex 
        \belowrulesep=0ex 
        \resizebox{\linewidth}{!}{
        \Huge
        \begin{tabular}{l | c c c}
            \toprule\rule{0pt}{1.1EM}
             & AP & AP$_{50}$ & AP$_{75}$\\
            \midrule\rule{0pt}{1.1EM}
            Scale Predictor & 34.2 & 53.5 & 33.2 \\
            + Scale Loss \& Fixed $\alpha / \beta$ & 33.9 & 53.2 & 35.0 \\
            + Learnable $\alpha / \beta$ & \multicolumn{3}{c}{Failed to converge} \\
            \rowcolor{LightGray} + Distribution Loss & \textbf{45.6$_{\color{MidnightBlue}\text{+11.4}}$} & 66.1 & 48.5\\
            \bottomrule
        \end{tabular}
        }
    \vspace*{0mm}
    \end{minipage}\hfill
    \begin{minipage}[b][][b]{0.65\linewidth}
    \vspace*{0mm}
    Note that all ablations are based on Ciliary-DETR-S-R50 with 20 epochs.
    As shown in \cref{tab:eff_losses}, the proposed losses yield a +11.4 AP improvement over the architecture-only design.
    Without the distribution loss, the model either degrades in performance or fails to converge, highlighting the importance of optimizing anchor sizes for stable scaling.
    \vspace*{0mm}
    \end{minipage}
\end{table}

\begin{table}[ht!]
    \vspace{-22pt}
    \begin{minipage}[b][][b]{0.33\linewidth} 
        \vspace*{0mm}
        \caption{Training Separation}
        \label{tab:comp_trn}
        \renewcommand*\arraystretch{1.4}
        \aboverulesep=0ex 
        \belowrulesep=0ex 
        \resizebox{\linewidth}{!}{
        \footnotesize
        \begin{tabular}{l | c  | c}
            \toprule\rule{0pt}{1.1EM}
             & AP & \# Epochs\\
            \midrule\rule{0pt}{1.1EM}
            Joint Training &  44.4 & 60 \\
            \rowcolor{LightGray} Separate Training & \textbf{45.6$_{\color{MidnightBlue}\text{+1.2}}$} & 40$^b$ + 20$^f$ \\
            \bottomrule
        \end{tabular}
        }
        \vspace*{0mm}
    \end{minipage}%
    \hfill
    \begin{minipage}[b][][b]{0.65\linewidth}
        \vspace*{0mm}
        \cref{tab:comp_trn} shows that our training strategy achieves a +1.2 AP improvement over fully joint training.
        This demonstrates the benefit of our single-pass multi-scale reasoning, which leverages robustness from MS augmentation.
        \vspace*{0mm}
    \end{minipage}
\end{table}

\begin{table}[ht!]
    \vspace{-22pt}
    \begin{minipage}[t][][b]{0.33\linewidth} 
        \caption{Distribution Types}
        \label{tab:eff_dist_type}
        \vspace*{0mm}
        \renewcommand*\arraystretch{1.5}
            \aboverulesep=0ex 
            \belowrulesep=0ex 
            \resizebox{\linewidth}{!}{
                \begin{tabular}{l | c c}
                    \toprule\rule{0pt}{1.1EM}
                     & AP & Final anchors $\mathcal{A}$ \\
                    \midrule\rule{0pt}{1.1EM}
                    Gaussian & Failed to learn & [0.00, 0.00] \\
                    Laplace & 43.1 & [0.00, 0.17] \\
                    \rowcolor{LightGray} Beta & \textbf{45.6} & [0.17, 0.68] \\
                    \bottomrule
                \end{tabular}
             }
        \vspace*{0mm}
    \end{minipage}%
    \hfill
    \begin{minipage}[t][][b]{0.65\linewidth}
        \vspace{9pt}
        As shown in \cref{tab:eff_dist_type}, Gaussian and Laplace distributions produce invalid or overly narrow estimates.
        This suggests that they fail to model network behavior effectively over the range [0, 1], whereas the Beta distribution captures the underlying trends and achieves the highest AP.
        \vspace*{0mm}
    \end{minipage}
    \vspace{-5pt}
\end{table}

\begin{table}[ht!]
    \vspace{-18pt}
    \begin{minipage}[b][][b]{0.33\linewidth} 
        \vspace*{0mm}
        \caption{Anchor Scaling}
        \label{tab:eff_offset}
        \renewcommand*\arraystretch{1.8}
        \aboverulesep=0ex 
        \belowrulesep=0ex 
        \resizebox{\linewidth}{!}{
            \Huge
            \begin{tabular}{l | c c c c}
                \toprule\rule{0pt}{1.1EM}
                 & AP & AP$_s$ & AP$_m$ & AP$_l$ \\
                \midrule\rule{0pt}{1.1EM}
                Naive Anchors & 45.3 & 24.1 & 49.2 & 65.0 \\
                \rowcolor{LightGray} $+$ $\xi$-driven Scaling & \textbf{45.6$_{\color{MidnightBlue}\text{+0.3}}$} & \textbf{25.6$_{\color{MidnightBlue}\text{+1.5}}$} & \textbf{49.9$_{\color{MidnightBlue}\text{+0.7}}$} & 64.6$_{\color{Maroon}\text{-0.4}}$ \\
                \bottomrule
            \end{tabular}
         }
        \vspace*{0mm}
    \end{minipage}%
    \hfill
    \begin{minipage}[b][][b]{0.65\linewidth}
        \vspace*{0mm}
        As presented in \cref{tab:eff_offset}, $\xi$-driven anchor scaling provides an additional +0.3 AP improvement.  
        The gain is particularly pronounced for small objects (+1.5 AP), while large objects exhibit a moderate 0.4 AP drop.
        \vspace*{0mm}
    \end{minipage}
    \vspace{-5pt}
\end{table}

\begin{table}[ht!]
    \vspace{-10pt}
    \begin{minipage}[b][][b]{0.28\linewidth} 
        \vspace*{0mm}
        \footnotesize
        \caption{Encoder Layers}
        \label{tab:eff_enc}
        \centering
        \renewcommand*\arraystretch{1.45}
        \aboverulesep=0ex 
        \belowrulesep=0ex 
        \resizebox{\linewidth}{!}{
            \begin{tabular}{l | c c c}
                \toprule\rule{0pt}{1.1EM}
                \# Enc. & AP & AP$_{50}$ & AP$_{75}$\\
                \midrule\rule{0pt}{1.1EM}
                0 & 44.8 & 65.3 & 47.9 \\
                \rowcolor{LightGray} 3 & \textbf{45.6$_{\color{MidnightBlue}\text{+0.8}}$} & 66.1 & 48.5 \\
                \bottomrule
            \end{tabular}
        }
        \vspace*{0mm}
    \end{minipage}
    \hfill
    \begin{minipage}[b][][b]{0.7\linewidth}
        \vspace*{0mm}
        \cref{tab:eff_enc} highlights the importance of architectural adaptability for accurate scaling behavior.
        The network without encoder layers yields 0.8 AP lower performance than the 3-layer encoder, which achieves only +0.2 AP over the baseline of [480, 800].
        \vspace*{0mm}
    \end{minipage}
    \vspace{-5pt}
\end{table}

\section{Conclusion}
We present a framework for adaptive image scaling that treats input resolution as a learnable variable rather than a fixed hyperparameter.
Our approach enables dynamic test-time adaptation through data-driven optimization objectives.
This work provides an orthogonal direction to architectural optimization, offering a new perspective on adaptive network optimization.



\medskip
{
\bibliographystyle{plain}
\bibliography{main}
}


\appendix
\clearpage

\renewcommand*{\thesection}{\Alph{section}}

\section{Discussion} 
\label{sec:disc}

\subsection{Relation with Visual Accommodation}
\noindent\textbf{Similarity and Difference.}
We provide an intuitive analogy between image scale and the behavior of the human eye lens, as both influence the perceived visual scale.
While these two mechanisms provide similar effect, they exhibit different characteristic in terms of information acquisition.
The human eye can capture additional visual details when focusing on real-world scenes, whereas changing image scale does not introduce new information.
Since the image is already captured and fixed by the camera, image scaling simply redistributes the same visual content over a different number of pixels.
In other words, the human eye can acquire new information from the physical world, whereas image scaling only alters how the observed data is sampled.

\noindent\textbf{Benefits from Input-level Adjustment.}
In the human visual system, accommodation enables (1) adaptive focusing, (2) efficient observation, and (3) a stable viewing experience without modifying the visual system itself.
Our method exhibits analogous properties:

\begin{enumerate}[leftmargin=*]
    \item[{(1)}] 
    \textit{Flexible internal prediction}: 
    Our framework models scale adaptation within the detection pipeline using a scale predictor that outputs a single image-level scale factor.
    As shown in \cref{fig:sf_plot}, this module enables data-driven scaling over a wide range of object sizes.
    
    \item[{(2)}] 
    \textit{Efficient Observation}: 
    As shown in \cref{tab:acc}, our approach improves both accuracy and computational efficiency, achieving approximately 18\% GFLOPs reduction compared to fixed-resolution inference.
    Over a wide scale range, the network also achieves consistent AP gains, demonstrating the effectiveness of adaptive input scaling.
    
    \item[{(3)}] 
    \textit{Robust performance}: 
    Our framework naturally incorporates multi-scale (MS) augmentation, enabling robust reasoning across scales.
    Results in \cref{tab:acc}, \cref{fig:iou_comp_baseline} and \cref{tab:cls_comp} demonstrate improved robustness to scale variation under a single inference budget.
    
    \item[{(4)}] 
    \textit{Modularity of input scaling}: 
    By introducing a scale predictor analogous to the ciliary muscle, we decouple input-level adaptation from the task-specific prediction pipeline.
    This preserves the underlying detection network while enabling direct control over input scaling, which also allows flexible adjustment of the computational budget.
    Our framework readily supports extensions such as MS augmentation, test-time scale variation (\cref{fig:test-time_res_variation}), and zero-shot scaling (\cref{fig:test-time_res_variation,tab:zeroshot_other_detectors}).
\end{enumerate}

\subsection{Limitations \& Future Work}

\begin{table}[t]
    \centering
    \caption{
        Results using DINO-DETR-R50~\citep{zhang2022dino}.
        This implementation follows the 24-epoch training schedule of DINO-DETR, with a learning rate decay at epoch 8 during joint training (denoted as $\ddagger$).
        Further optimization under this setting may lead to improved performance.
    }
    \label{tab:dino_impl}
    \renewcommand*\arraystretch{1.6}
    \aboverulesep=0ex 
    \belowrulesep=0ex 
    \resizebox{\linewidth}{!}{
        \begin{tabular}{l | c | c | c c}
            \toprule\rule{0pt}{1.1EM}
              & Epochs & Trained Detector & AP & GFLOPs \\
            \midrule\rule{0pt}{1.1EM}
            DINO-DETR-4scale-R50 & 24 & \ding{51} & 50.6 & 246 \\
            \rowcolor{UltraLightGray} \texttt{+ Ciliary-DETR-Zeroshot / Ciliary-DETR-Zeroshot-DINO-4scale-R50} & 24 & \ding{51} & 50.5 & \textbf{202} \\
            \rowcolor{UltraLightGray} \texttt{+ Joint Training / Ciliary-DETR-DINO-4scale-R50} & 12$^\dagger$ + 12$^\ddagger$ & \ding{55} & 50.3 & \textbf{208} \\
            \bottomrule
        \end{tabular}
    }
\end{table}

\noindent\textbf{Generalization over Detection Architectures.}
The proposed scale predictor is designed to be architecture-agnostic and demonstrates promising generalization across different detectors.
As shown in \cref{tab:zeroshot_other_detectors}, zero-shot application achieves performance comparable to fixed-resolution inference while reducing computational cost by 17–18\%.
Additional results in \cref{tab:dino_impl} further demonstrates its applicability under joint training.
While these results are encouraging, a more comprehensive evaluation across diverse architectures remains future work.

\noindent\textbf{Annotation Dependency.}
Our method relies on object size information derived from ground-truth annotations.
In scenarios where such annotations are unavailable, this requirement may limit applicability.
Such information can be approximated using pretrained models that provide saliency~\citep{caron2021emerging, oquab2023dinov2, simeoni2025dinov3} or segmentation cues~\citep{wang2023cut, li2024promerge}.
Extending the framework to operate without explicit annotations is left for future work.

\noindent\textbf{Optimization for Anchor Sizes.}
The current formulation estimates anchor sizes through a multi-stage procedure, and this can be optimized in several directions.
(1) Optimizing anchor sizes without explicit global representation may enable an accurate anchor estimation with only a single network representation.
This can be achieved by incorporating the input scale range (i.e., $\tau_{min}$ and $\tau_{max}$) into the parametric distribution, introducing explicit modeling of network-level scaling capacity.
(2) More principled approaches, such as quantile-based estimation, Bayesian inference, or EM-style optimization, could further improve robustness and accuracy.

\noindent\textbf{Integration with Token-level Pruning.}
One promising direction is the integration with token-level pruning~\citep{zheng2023less, hou2024salience}, which can reduce the computational cost of high-resolution inputs.
Such pruning methods retain informative regions while removing background or noisy tokens, enabling more effective utilization of image information.
Since our work is orthogonal to such architectural modifications, combining image scaling with token pruning may further improve the efficiency of high-resolution processing, which we leave for future work.

\noindent\textbf{Implementation Aspects.}
The current scale prediction uses a bounded formulation, which may not fully utilize the available output range.
A reparameterization such as $\phi = \phi_{\sigma} \cdot (\tau_{max} - \tau_{min}) + \tau_{min}$ could improve numerical stability and coverage.
In addition, the use of bilinear interpolation for resizing may introduce information loss.
The use of learnable or higher-fidelity resampling methods could mitigate such limitation.
A systematic investigation of these design choices is left for future work.

\subsection{Supernet-based Resolution Optimization} 
\label{ssec:more_resolution_info}

\noindent\textbf{Association between Supernet and MS training.}
Supernets were originally introduced in neural architecture search (NAS)~\cite{guo2020single, cai2019once, li2020gan} to enable efficient exploration of multiple architectural configurations within a single over-parameterized network.
Such networks are typically trained with stochastic sampling strategies, after which sub-networks (i.e., subnets) are extracted to satisfy specific constraints (e.g., FLOPs or latency).
Motivated by this perspective, we draw an analogy between supernet training and multi-scale (MS) training in object detection.
Specifically, we interpret a detector trained with MS augmentation as an implicit supernet that spans a range of input resolutions.
Accordingly, different resolution configurations can be viewed as subnets, each corresponding to a different computational budget.
During joint training, we vary the resolution range in a manner analogous to subnet sampling, enabling effective exploration of subnet-like models with different resolution budgets.

\noindent\textbf{Resolution Configuration.}
We formalize the resolution space using a configuration set $\Upsilon$.
The configurations for supernet-like MS augmentation and subnet-like joint training are defined as,
\begin{equation}
    \begin{split}
        \Upsilon_{super}: \mathbb{C}(\tau_{min}, \tau_{max}, \rho), \quad
        \Upsilon_{k}: \mathbb{C}(\tau_{min}, \tau^{k}_{max}, \rho),
    \end{split}
\end{equation}
where $\Upsilon_{k} \subseteq \Upsilon_{super}$ and $\tau_{max}^k\le \tau_{max}$.
Here, $\tau_{min}$ and $\tau_{max}$ denote the minimum and maximum scale factors, and $\rho$ represents the resolution step size.
The configuration function $\mathbb{C}(\cdot)$ generates a discrete set of resolutions as,
\begin{equation}
    \begin{split}
    &\mathbb{C}(\tau_{min}, \tau_{max}, \rho) = [(W_0, H_0), ~(W_0 +\rho, H_0 + \rho), (W_0+2\rho, H_0+2\rho),~...,~(W + n\rho, H + n\rho)],   
    \end{split}
\end{equation}
where $(W_0, H_0)$ denotes the minimum resolution determined by $\tau_{min}$, and $n$ is the largest integer such that the resulting resolution does not exceed the maximum scale defined by $\tau_{max}$.

\section{More Analysis \& Experiments} 
\label{sec:more_expr}

\begin{table}[t]
    \centering
    \caption{
        Implementation of MS TTA by progressively adding components to the baseline network.
        We select the top-k bounding boxes based on classification scores, following the maximum number of predictions (300) in DN-DETR.
        When both selection and NMS are used, we first perform NMS and then apply top-$k$ selection to maintain a fixed number of outputs.
    }
    \label{tab:ms_tta_impl}
    \renewcommand*\arraystretch{1.6}
    \aboverulesep=0ex 
    \belowrulesep=0ex 
    \resizebox{.95\linewidth}{!}{
        \begin{tabular}{l | l | c | c || c}
            \toprule\rule{0pt}{1.1EM}
              & Specification & Resolution & AP (R50) & AP (Swin-S) \\
            \midrule\rule{0pt}{1.1EM}
            MS-Wide-Baseline & - & 1280 & 47.2 & 50.5\\\hline
             + \texttt{MS TTA} & \texttt{image-level top-k (Score$_\text{class}$, k=300)} & & 9.9 & 10.2 \\
             & \texttt{+ class-level NMS (IoU threshold: 0.5)} & & 45.2 & 47.8 \\
             & \texttt{+ increasing threshold: 0.6} & & 45.4 & 48.0 \\
             & \texttt{+ increasing threshold: 0.7} & \multirow{-4}{*}{[320, 1280]} & 44.8 & 47.3 \\
             \cdashline{2-5}
             & \texttt{image-level top-k (Score$_\text{class}$, k=300)} &  & - & - \\
             & \texttt{+ increasing threshold: 0.6} & \multirow{-2}{*}{[320, 2336]} & 46.1 & 49.0 \\
             \cdashline{2-5}
             & \texttt{image-level top-k (Score$_\text{class}$, k=300)} &  & - & - \\
             & \texttt{+ increasing threshold: 0.6} & \multirow{-2}{*}{[800, 2336]} & 47.5 & 50.8 \\
            \bottomrule
        \end{tabular}
    }
\end{table}



\subsection{More Analysis of Test-time Multi-Scale Strategies} 
\label{ssec:more_scale_selection_comp}
\cref{tab:comp_with_mstrn} compares different test-time multi-scale strategies on trained baseline networks.
We analyze each method as follows.

\begin{enumerate}[leftmargin=*]
\item[(1)] Random scale selection: 
\cref{tab:comp_with_mstrn} shows that the network does not adapt well to random scale selection, although such randomness is beneficial during training.
This suggests that test-time resolution should be selected in a data-dependent manner.

\item[(2)] MS TTA (multi-scale test-time augmentation):
After merging predictions, we select bounding boxes using top-$k$ selection based on classification scores, followed by NMS\footnote{
In DETR-based models, ground-truth and predictions are matched using the Hungarian algorithm during training, but there is no explicit mechanism for filtering overlapping predictions.
Therefore, we introduce a simple NMS, as there is limited prior work on aggregating predictions across scales.
}.
When NMS is applied, performance improves significantly, but remains lower than that of the baseline network.
This indicates that while filtering is important, more principled methods for TTA in DETR frameworks are needed.
In contrast, our method avoids such merging and filtering steps, demonstrating the practical effectiveness of single-pass scaling.

\item[(3)] BFS (brute-force search):
Following DETR, BFS is implemented using Hungarian matching to evaluate predictions against ground-truth annotations, considering both classification and localization quality.
We select a single resolution based on the average IoU of matched predictions, which serves as an approximate upper bound on performance for single-resolution selection.
Our method achieves results close to this optimal selection, with only a 0.4--0.5 AP gap.
\end{enumerate}

Additionally, our method improves AP even within the training resolution range of the baseline, as shown in \cref{fig:test-time_res_variation}, indicating the effectiveness of data-driven resolution scaling.

\begin{table}[t]
    \caption{
        Latency measurement of detection networks on an RTX 3090 GPU. 
        We average results over 3 runs, each with 2000 images, resulting in 6000 total inference iterations.
        Image pre-loading refers to loading images onto the GPU before executing the network.
    }
    \label{tab:latency_measurement}
    \centering
    \renewcommand*\arraystretch{1.6}
    \aboverulesep=0ex 
    \belowrulesep=0ex 
    \resizebox{.95\linewidth}{!}{
        \begin{tabular}{l | c c  c | c c  | c c }
        \toprule\rule{0pt}{1.1EM}
         & & & & \multicolumn{2}{c|}{R50} & \multicolumn{2}{c}{Swin-Small} \\\cmidrule{5-8}
         & \multirow{-2}{*}{Strategy} & \multirow{-2}{*}{Resolution} & \multirow{-2}{*}{Image Pre-loading} & GFLOPs & Latency & GFLOPs & Latency \\
        \midrule\rule{0pt}{1.1EM}
         &  & & \ding{55} &  & 37ms  & & 61ms \\
        \multirow{-2}{*}{MS baseline}  & \multirow{-2}{*}{Fixed} & \multirow{-2}{*}{800} & \ding{51} & \multirow{-2}{*}{90} & 38ms & \multirow{-2}{*}{180} & 61ms \\
        \rowcolor{UltraLightGray} Ciliary-DETR & Off-line Adaptive Scaling & [480, 800] & \ding{51} & 74 & \textbf{34ms} & 148 & \textbf{58ms} \\
        \bottomrule
    \end{tabular}
     }
\end{table}

\subsection{Prediction Latency}
\cref{tab:latency_measurement} shows a latency comparison between the MS baseline and Ciliary-DETR-R50.
This result suggests that our adaptive scaling improves latency efficiency.
Specifically, we adopt an offline scale predictor that determines the appropriate scale before running the detection network, while the detection network processes only the scaled input at inference time.
This design decouples data pre-processing overhead from network inference time.
For a fair comparison, we also enable image pre-loading on the GPU for the MS baseline, to match our pipeline, where the scale predictor outputs GPU-resident tensors.
This setting slightly increases the measured latency due to reduced available GPU resources.
Nevertheless, our method still outperforms the baseline even when pre-loading is enabled, demonstrating its effectiveness in practical inference scenarios.

\begin{figure}[t]
    \centering
    \includegraphics[width=\linewidth]{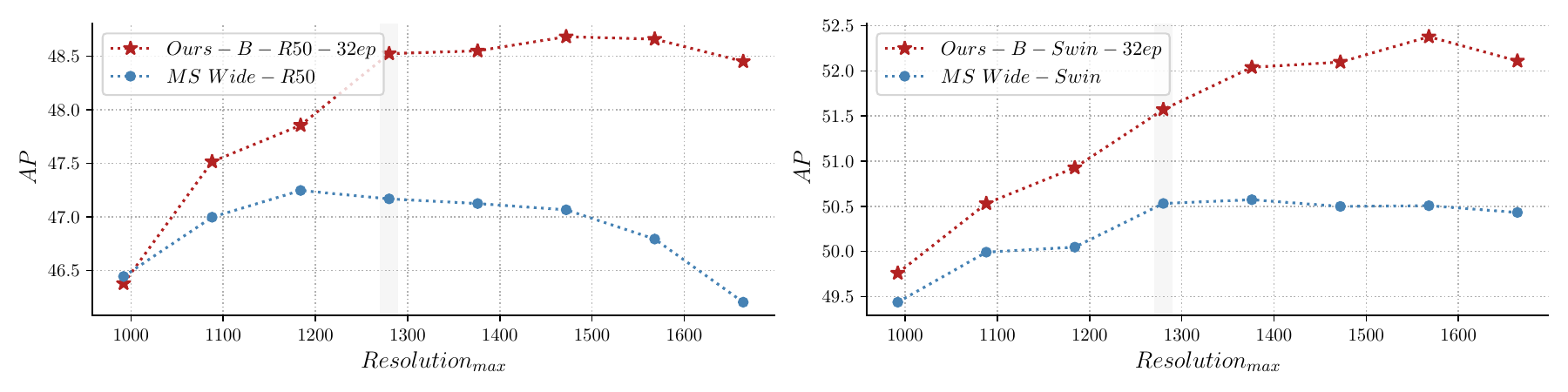}
    \caption{
        (Left) The same results as \cref{fig:test-time_res_variation}.
        (Right) Test-time robustness of networks based on the Swin-Small backbone.
    }
    \label{fig:res_variation_all}
\end{figure}

\begin{table}[t]
    \caption{
    Effect of image resizing on the baseline with a Swin-Small backbone.
    Second Up-sampling refers to an additional up-sampling process after the dataloading.
    For the MS baseline with only 2nd Up-sampling, we perform image up-scaling to a resolution of 1280 using an initial resolution of 600, following the implementation of our networks.
    }
    \centering
    \label{tab:baseline_swin_sp_use}
    \renewcommand*\arraystretch{1.6}
    \aboverulesep=0ex 
    \belowrulesep=0ex 
    \resizebox{\linewidth}{!}{
        \begin{tabular}{l | c c | c | c c }
        \toprule\rule{0pt}{1.1EM}
         & Our Adaptive Scaling & 2nd Up-sampling & Up-sampling Strategy & AP & GFLOPs \\\hline
        MS-Wide Baseline (Swin-Small) & \ding{55} & \ding{55}& \ding{55} & 50.5 & 450 \\
         & \ding{55} & \ding{51} & Our Custom Interpolation & 49.7 & - \\
         & \ding{55} & \ding{51} & \texttt{Pytorch}'s Bicubic Interpolation & 49.7 & - \\
         \rowcolor{DarkGray} \texttt{+ Ciliary-DETR-Base-Zeroshot} & \ding{51} & \ding{51} & Our Custom Interpolation & 49.8 & 374 \\
         \rowcolor{DarkGray} & \ding{51} & \ding{51} & \texttt{Pytorch}'s Bicubic Interpolation & \textbf{49.9} & - \\ \hline\hline
         MS-Wide Baseline (R50) & \ding{55} & \ding{55}& \ding{55} & 47.2 & - \\
          & \ding{55} & \ding{51} & Our Custom Interpolation & 47.2 & - \\
        \bottomrule
    \end{tabular}
     }
\end{table}

\subsection{More Experiments for Test-time Robustness against Resolution Variation}
\noindent\textbf{Swin-based Networks.}
\cref{fig:res_variation_all} demonstrates strong test-time robustness for networks with Swin backbones.
Similar to the ResNet-based results, our adaptation scheme consistently outperforms the baseline across varying resolutions.
When evaluated at a resolution of 1568, our network achieves the largest gain of +0.8 AP (51.6 $\rightarrow$ \textbf{52.4} AP) compared to the training resolution of 1280.
Compared to the baseline, our network achieves the largest gain of +1.9 AP at a resolution of 1568, further highlighting the test-time robustness of the proposed adaptation mechanism.

\noindent\textbf{Effect of Backbone Networks.}
Interestingly, the MS baseline with Swin exhibits stronger extrapolation capability to unseen resolutions than its ResNet counterparts.
The Swin-based baseline achieves a slight AP decrease of 0.1 at 1664, whereas the baseline with R50 shows -1.0 AP.
This difference may stem from the architectural characteristics: attention-based models such as Swin may inherently exhibit greater robustness to resolution variation compared to purely convolutional networks.

\noindent\textbf{Resizing Sensitivity in Swin-based Networks.}
\cref{tab:baseline_swin_sp_use} shows the test-time sensitivity of the Swin-based network to resizing operations. 
These results indicate low robustness to double interpolation in Swin backbone, unlike ResNet-50.
Following Park \textit{et al.}~\cite{park2022vision}, multi-head self-attention (MSA) can be interpreted as a low-pass filter, whereas convolution is viewed as a high-pass filter.
This interpretation suggests that the Swin network may naturally be vulnerable to low-frequency noise introduced by interpolation. 
This effect can be mitigated by training, as demonstrated in \cref{tab:acc}, suggesting the importance of joint training for adaptive scaling in Swin-based networks.

\noindent\textbf{Zero-shot Scaling for Swin-based Network.}
As shown in \cref{tab:baseline_swin_sp_use}, the performance of this resizing-sensitive network improves by +0.1 AP with a 17\% reduction in GFLOPs when applying zero-shot adaptive scaling.
Furthermore, naive bicubic interpolation yields an additional +0.1 AP gain, indicating that the network is sensitive to the choice of resizing operations.

\begin{table}[t]
    \caption{
        More results for zero-shot scaling.
        Scaling range is [480, 800] for Ciliary-DETR-Zeroshot and [320, 864] for Ciliary-DETR-S-Zeroshot.
    }
    \label{tab:more_zeroshot}
    \centering
    \renewcommand*\arraystretch{1.6}
    \aboverulesep=0ex 
    \belowrulesep=0ex 
    \resizebox{.8\linewidth}{!}{
        \begin{tabular}{l | c | c c | >{\columncolor{UltraLightGray}}c >{\columncolor{UltraLightGray}}c | >{\columncolor{LightGray}}c >{\columncolor{LightGray}}c}
        \toprule\rule{0pt}{1.1EM}
         & & \multicolumn{2}{c|}{\texttt{Max Inference}} & \multicolumn{2}{c|}{\cellcolor{UltraLightGray}\texttt{Ours-Zeroshot}} & 
        \multicolumn{2}{c}{\cellcolor{LightGray}\texttt{Ours-S-Zeroshot}} \\ \cmidrule{3-8}
         & \multirow{-2}{*}{Epochs} & AP & GFLOPs & AP & GFLOPs & AP & GFLOPs \\\hline
         Deform-DETR-R50 & 50 & 47.0 & 172 & 47.0 & \textbf{141} & \textbf{47.3} & \textbf{166} \\\hline
         DINO-DETR-4scale-R50 & 12 & 49.2 & 246 & \textbf{49.4} & \textbf{202} & \textbf{49.7} & \textbf{231} \\
          & 24 & 50.6 & 246 & 50.5 & \textbf{202} & \textbf{50.7} & \textbf{231} \\\hline
         H-DETR-R50 & 12 & 49.1 & 236 & 49.1 & \textbf{192} & \textbf{49.6} & \textbf{221} \\
         & 36 & 50.3 & 236 & 50.3 & \textbf{192} & 50.3 & \textbf{221} \\
        \bottomrule
    \end{tabular}
     }
\end{table}

\subsection{More Utilization in Zero-shot Scaling}
\cref{tab:more_zeroshot} demonstrates results from zero-shot scaling with ours-S under [320, 864].
Notably, the performance of networks consistently improves in a zero-shot setting, especially for networks with 12 training epochs.
Our zero-shot scaling on DINO-DETR and H-DETR both achieve +0.5 AP, while networks with longer epochs (e.g., H-DETR with 36 epochs) show limited performance gains.
This result suggests that our adaptive scaling, which includes unseen resolutions (e.g., [320, 480) and (800, 864]), is beneficial for improving detection performance for networks with lower convergence.

\noindent\textbf{Conceptual Analysis for Performance Efficiency.}
Generally, networks with multiple features tend to be more robust to input scale variations compared to single-scale networks.
Following the anchor size formulation, the scale-dependent performance of such networks can be approximated as,
\begin{equation}
    \mathbb{D}_{multi} (s; s^m_{min}, s^m_{max} ), 
    \quad
    \mathbb{D}_{single} (s; s^s_{min}, s^s_{max} ), 
    \quad
    s^m_{min} \geq s^s_{min} \text{ and } s^m_{max} \geq s^s_{max},
\end{equation}
where $\mathbb{D}_*$ denotes the detection network, and $s^*_{min}$ and $s^*_{max}$ denote the saturation sizes of the detection performance.
Thus, the networks with robust performance can effectively utilize a scale predictor trained on single-scale networks.

\section{More Implementation Details} 
\label{sec:more_impl_details}

\noindent\textbf{Training Environment.} 
Our models are trained on servers equipped with Nvidia A6000 GPUs, specifically, 
\begin{itemize}[leftmargin=*]
    \item Ours-R50/Ours-S-R50/Ours-M-R50 and Ours-Swin-S: 2 GPUs
    \item Baseline-Swin-S/Ours-B-Swin-S: 8 GPUs
    \item All other models: 4 GPUs
\end{itemize}
We follow the training configuration of DN-DETR, which adopts a learning rate (LR) of 1e-4 with the Adam optimizer and a batch size of 16.
The baseline DN-DETR is trained for 50 epochs, consisting of 40 epochs without LR decay and 10 epochs with a LR decay of 0.1$\times$.
When training our networks, supernet-like MS training is performed during 40 epochs as the baseline epochs without LR decay.
We fine-tune the detection networks with the scale predictor during 10, 20, or 32 epochs, while LR decaying is applied after 4, 15, and 24 epochs, respectively.
The proposed losses are a weighted sum with existing losses using $\lambda_{scale}=5$ and $\lambda_{dist}=1$.

\noindent\textbf{Image Resolution.} 
In our experiments, the image resolution is adjusted while preserving the aspect ratio of the image between width and height.
This strategy resizes the shorter edge to the target resolution, and the longer edge is constrained by defining a maximum value.
In the baseline network, the maximum is set as 1333 when the shorter edge is 800; for example, resolution 800 may correspond to 800×1000 or 800×896, and the longer size must not exceed 1333.
We set the initial image resolution as 6000, with the corresponding maximum of 1000, proportional to this baseline setting.
The scale predictor operates at a down-sampled resolution of 240, which is resized by 0.4$\times$.
We resize the image from the initial resolution according to the obtained scale factor.
We discretize the resolution by divisibility of $\rho$ through numerical rounding as, $f_{disc}(x, \rho)=round(x/\rho) \cdot \rho$.
We also implement a bilinear interpolation that receives the desired size as differentiable tensors, which is detailed in \cref{tab:eff_int}.

\begin{table*}[t]
\begin{minipage}[b][][b]{0.55\linewidth}
    \centering
    \caption{
        Comparison between the implementation of bilinear interpolation (denoted as B.I.).
        We compare the inference result with different interpolation layers.
    }
        \label{tab:eff_int}
        \renewcommand*\arraystretch{1.6}
        \aboverulesep=0ex 
        \belowrulesep=0ex 
        \resizebox{\linewidth}{!}{
            \begin{tabular}{l | c c | c c }
                \toprule\rule{0pt}{1.1EM}
                & \multicolumn{2}{c|}{\texttt{Pytorch-based B.I.}} & \multicolumn{2}{c}{\texttt{Custom B.I.}} \\\cmidrule{2-5}
                & AP & GFLOPs & AP & GFLOPs \\
                \midrule\rule{0pt}{1.1EM}
                \texttt{Ciliary-DETR-S-R50} & 45.6 & 84 & 45.6 & 85 \\
                \texttt{Ciliary-DETR-M-R50} & 47.1 & 130 & 47.2 & 132 \\
                \texttt{Ciliary-DETR-B-R50} & 48.2 & 185 & 48.2 & 186 \\
                \bottomrule
            \end{tabular}
        }
\end{minipage}%
\hfill
\begin{minipage}[b][][b]{0.43\linewidth}
    \centering
    \caption{
        Methods for loss combination.
        Loss Weighting refers to adaptive loss-driven weighting for our losses.
    }
    \label{tab:eff_lambda}
    \small
    \renewcommand*\arraystretch{1.8}
    \aboverulesep=0ex 
    \belowrulesep=0ex 
    \resizebox{\linewidth}{!}{
        \begin{tabular}{l | c | c }
            \toprule\rule{0pt}{1.1EM}
            &  Loss Weighting & AP \\
            \midrule\rule{0pt}{1.1EM}
            $\lambda_{scale}=3$, $\lambda_{dist}=1$ & \ding{55} & 45.4 \\
            $\lambda_{scale}=5$, $\lambda_{dist}=1$ & \ding{55} & 45.5 \\
            $\lambda_{scale}=5$, $\lambda_{dist}=1$ & \ding{51} & \textbf{45.6} \\
            \bottomrule
        \end{tabular}
    }
\end{minipage}
\end{table*}

\noindent\textbf{Additional Information.}
To reduce computational cost, the scale predictor uses an intermediate result in the backbone network, specifically the third ResNet block.
In the scale loss, we select $\epsilon$ to 0.0025, which results in an $x$-values of [-6, 6].
For the distribution loss, we clip the localization loss using $\min(\mathcal{L}'_{det}, 1)$ to reduce the influence of outliers.
Although networks with [480, 800] do not directly follow our supernet-based training setup, we apply the anchor scaling to enable more aggressive up-scaling. 
The training environment of the LVIS follows the same configuration as COCO.

\noindent\textbf{Weight-driven Loss Aggregation.}
To stabilize the convergence of the scale predictor, we apply an additional weighting factor~\cite{lin2017focal} that regulates the impact of proposed losses.
We define this factor as the average scalar value of the existing losses, so that the adaptive image loss becomes more textitasized when the detector exhibits lower performance.
The weighting factor is denoted as the average of all existing loss terms as,
\begin{equation}
    \mathcal{L}_\text{total} = \frac{1}{V}\sum_v \mathcal{L}'_v \cdot (\lambda_\text{scale} \cdot \mathcal{L}_\text{scale} + \lambda_\text{dist} \cdot \mathcal{L}_\text{dist}) + \sum_v \lambda_v \cdot \mathcal{L}_v,
\end{equation}
where $\mathcal{L}'_*$ denotes the detached loss and $\lambda_*$ denotes the weight of each loss term.
\Cref{tab:eff_lambda} summarizes our experiments on this loss combination method.
Note that the weighting factor $\lambda_\text{avg}$ is obtained from the loss magnitude without gradient propagation.
This result suggests the effectiveness of adaptive loss weighting, which leads to +0.1 AP.

\begin{figure}
    \centering
    \includegraphics[width=\linewidth]{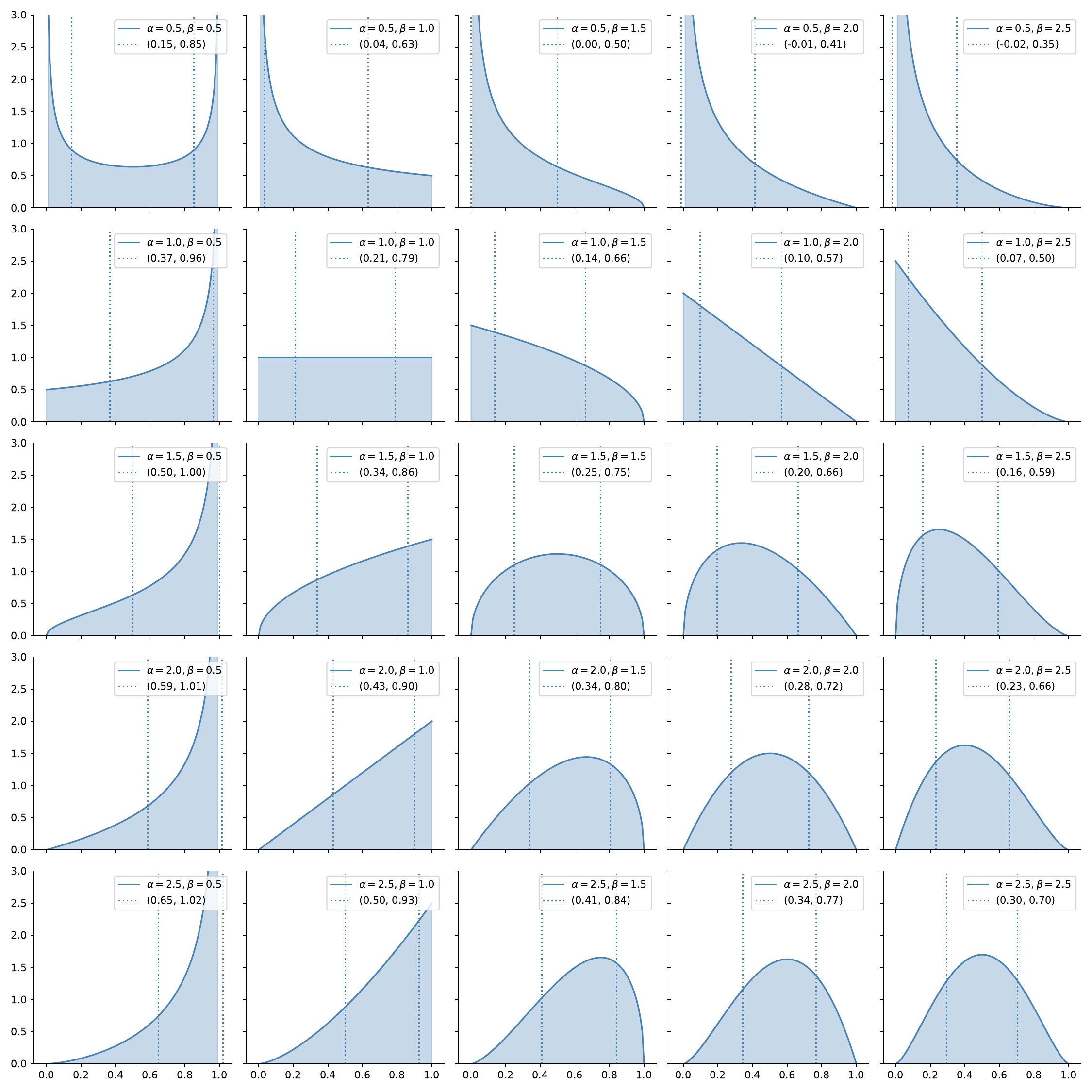}
    \caption{
        Example of Beta distributions with various $\alpha$, $\beta$. 
        The dashed line indicates $\mu \pm \sigma$.
        }
    \label{fig:beta_examples}
\end{figure}

\section{Formal Interpretation \& Proof}
\label{sec:formal}

\subsection{Convexity of Scale Loss}
\label{ssec:convex_scale_loss}

Let the basic function of scale loss be expressed as, 
\begin{equation} 
\label{eq:bce_func}
f_{sc}(\theta, a)=- \left( \theta \cdot log(a)+(1-\theta)\cdot log(1-a) \right),
\end{equation}
where $\theta \in [0,1]$ and $a \in (0,1)$, and the logarithm denotes the natural logarithm. 
To prove convexity, we examine the second derivative of $f_{sc}$ with respect to $a$:
\begin{equation}
    \frac{\partial^2{f_{sc}}}{\partial{a}^2}=\frac{\theta}{a^2} + \frac{1-\theta}{(1-a)^2}.
\end{equation}
Since both terms are nonnegative for all $a \in (0,1)$, it follows that
\begin{equation}
    \frac{\partial^2{f_{sc}}}{\partial{a}^2} \ge 0, \quad\forall{a} \in (0,~1),
\end{equation}
which implies that $f_{sc}$ is convex with respect to $a$ on its domain.
The final scale loss is obtained by averaging $f_{sc}$ over instances with nonnegative weights.
Since convexity is preserved under finite nonnegative weighted sums, the overall scale loss remains convex with respect to $a$.

\subsection{Anchor Sizes derived from Beta Distribution} 
\label{ssec: beta_distribution}

In \textit{distribution loss}, we aim to learn the performance tendency across scales using the Beta distribution.
The probability density function (PDF) of the Beta distribution is provided as,
\begin{equation}
    f_{beta}(x; \alpha, \beta)=\frac{x^{\alpha-1}(1-x)^{\beta-1}}{B(\alpha, \beta)},
\end{equation}
where $B(\alpha, \beta)$ denotes the incomplete Beta function.
These anchor sizes are induced from the Beta distribution using,
\begin{align}
     \mu \pm \sigma = \frac{\alpha}{\alpha+\beta} \pm \sqrt\frac{\alpha \beta}{(\alpha+\beta)^2(\alpha+\beta+1)},
\end{align}
where $\mu$ and $\sigma$ denote the mean and standard deviation, respectively.
Here, we aim to identify the range from possible $\alpha$ and $\beta$ that satisfy $[\mu-\sigma, \mu+\sigma] \in [0, 1]$.
We assume positive $\alpha$ and $\beta$ values.

i) 
For the upper bound, $\mu + \sigma$, the inequality to restrict $\leq 1$ can be derived as,
\begin{align}
    \frac{\alpha}{\alpha+\beta} + \sqrt\frac{\alpha \beta}{(\alpha+\beta)^2(\alpha+\beta+1)} \leq 1 \\
    \sqrt\frac{\alpha \beta}{(\alpha+\beta)^2(\alpha+\beta+1)} \leq 1  - \frac{\alpha}{\alpha+\beta}\\
    \sqrt\frac{\alpha \beta}{(\alpha+\beta)^2(\alpha+\beta+1)} \leq \frac{\beta}{\alpha+\beta} \\
    \frac{\alpha \beta}{(\alpha+\beta)^2(\alpha+\beta+1)} \leq \frac{\beta^2}{(\alpha+\beta)^2} \\
    \alpha \leq \beta (\alpha+\beta+1) \\
    \alpha(1-\beta) \leq \beta(1+\beta) \label{ineq:upper_bnd}
\end{align}

ii) For $\mu - \sigma$, the inequality for lower bounds $\geq 0$ is defined as,
\begin{equation}
  \mu - \sigma = \frac{\alpha}{\alpha+\beta} - \sqrt\frac{\alpha \beta}{(\alpha+\beta)^2(\alpha+\beta+1)} \geq 0.
\end{equation}
This leads to the simplified form represented as,
\begin{equation} \label{ineq:lower_bnd}
  \alpha (1+\alpha) \geq \beta (1-\alpha).
\end{equation}

\begin{figure*}[t]
    \centering
    \begin{subfigure}[l]{0.24\textwidth}
        \includegraphics[width=\linewidth]{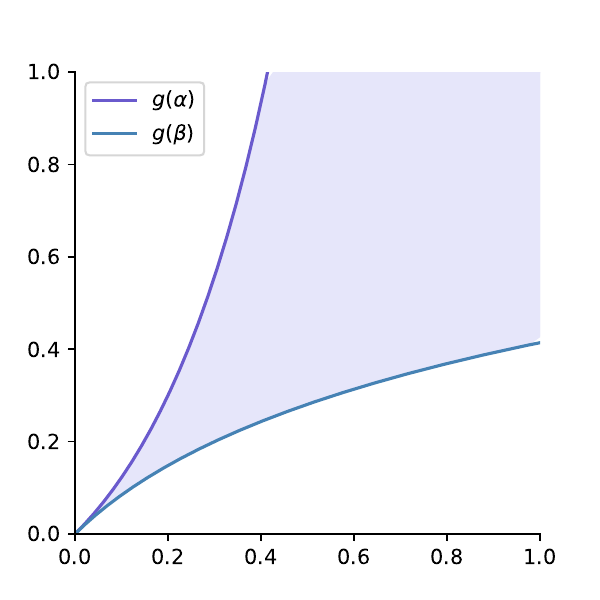}
        \caption{Inequality Visualization}
        \label{fig:vis_ineq}
    \end{subfigure}\hspace*{\fill}%
    \begin{subfigure}[r]{0.75\textwidth}
        \includegraphics[width=\linewidth, trim={0cm 0.4cm 0.2cm 0cm}, clip]{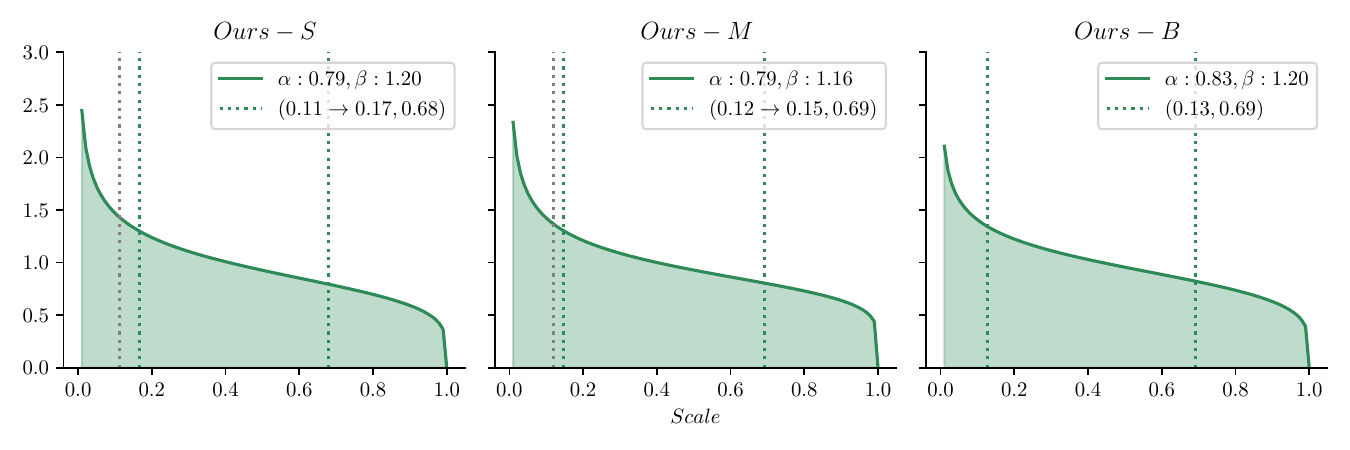}
        \caption{Obtained Boundaries}
        \label{fig:ours_beta}
    \end{subfigure}%
    \caption{
    (a) Visualization of possible $\alpha, \beta$ values when $\alpha, \beta<1$. 
    This graph indicates that $\alpha$ must be greater than 0.415 when $\beta = 1$ to satisfy the inequality.
    (b) Visualization of the obtained boundaries and the Beta distribution, which indicates similar results across different resolution ranges.
    The gray and green lines indicate the original boundaries from the distribution and the modified range by the lower boundary scaling, respectively.
    }
\end{figure*}

For $\beta > 1$ or $\alpha > 1$, \cref{ineq:upper_bnd} or \cref{ineq:lower_bnd} always hold, respectively.
When $\beta < 1$, \cref{ineq:upper_bnd} can be modified as,
\begin{equation} \label{ineq:lower_bnd_for_less_beta}
    \alpha \leq \frac{\beta (1+\beta)}{1-\beta},
\end{equation}
Also, the form of \cref{ineq:lower_bnd} is changed for $\alpha < 1$ as,
\begin{equation} \label{ineq:lower_bnd_for_less_alpha}
    \beta \leq \frac{\alpha (1+\alpha)}{1-\alpha}.
\end{equation}
The possible solutions from these two inequalities are illustrated in \cref{fig:vis_ineq}.
This figure suggests that the resulting anchor sizes remain [0, 1] when $\alpha$ and $\beta$ take moderate values (i.e., neither of them is extremely large or close to zero).
This indicates that a Beta distribution can reasonably model the network performance unless the detection network provides extreme scale-related accuracy imbalance (e.g., highly skewed accuracy such as 90\% vs. 10\% for large vs. small objects).

\begin{figure*}[t]
    \centering
    \includegraphics[width=\linewidth]{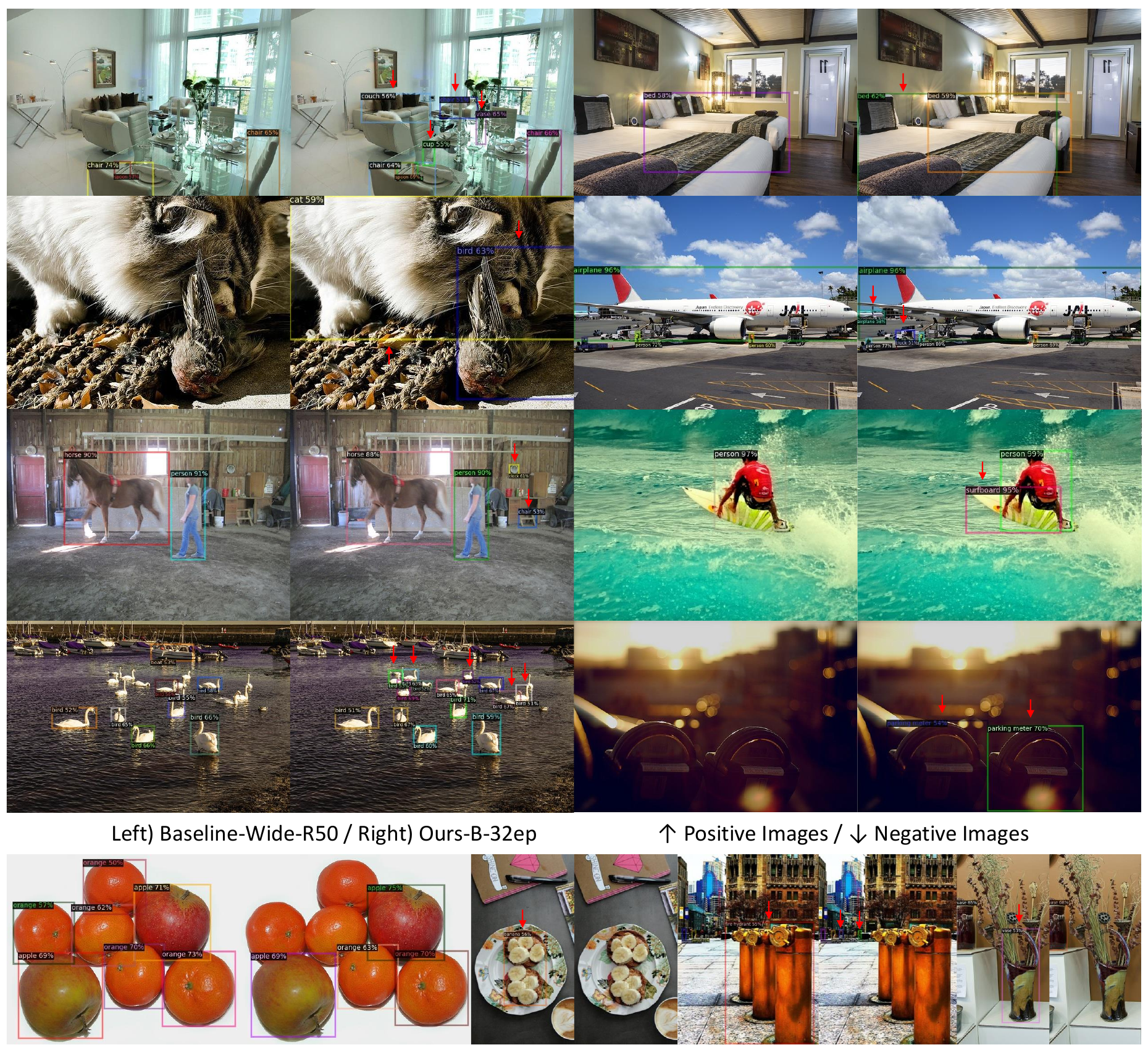}
    \caption{
        Image examples of COCO \texttt{val} that provide higher or lower performance than the baselines with ResNet-50 in \cref{tab:comp_with_mstrn}.
        The arrow indicates the difference in prediction results, implying that our method can provide robust performance at a wide range of scales, not only improving performance for specific object scales.
        Specifically, images with extremely high or various object scales benefit more from our adaptive method, leading to more accurate prediction performance.
    }
    \label{fig:res_more_img_examples}
\end{figure*}

\begin{figure*}[t]
    \centering
    \includegraphics[width=\linewidth]{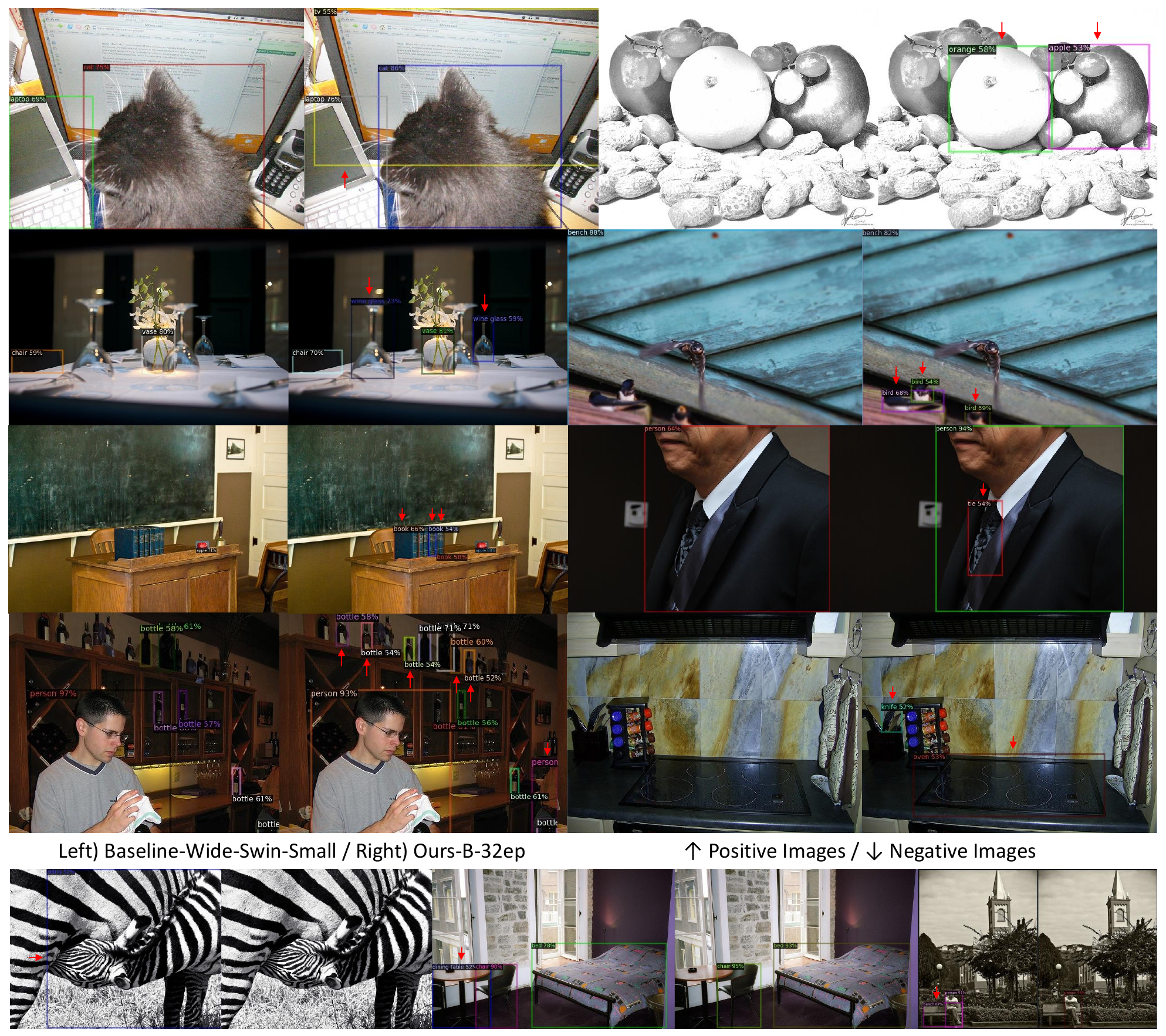}
    \caption{Image examples of COCO \texttt{val} that provide higher or lower performance than the baselines with Swin-Small in table 3.}
    \label{fig:swin_more_img_examples}
\end{figure*}


\end{document}